\def\paperlanguage{}
\pdfoutput=1 

\documentclass[letterpaper, 10 pt, conference]{ieeeconf}  

\usepackage{bm}
\usepackage{cite}
\usepackage{flushend}
\include{preamble}

\IEEEoverridecommandlockouts                              

\title{\LARGE \textbf
  {
    \switchlanguage%
    {%
      Object Recognition, Dynamic Contact Simulation, Detection, and Control of the Flexible Musculoskeletal Hand\\ Using a Recurrent Neural Network with Parametric Bias
    }%
    {%
      パラメトリックバイアスを含む再帰的ニューラルネットワークを用いた筋骨格柔軟ハンドの物体認識, 動的シミュレーション・接触検知・制御手法の開発
    }%
  }
}

\author{Kento Kawaharazuka$^{1}$, Kei Tsuzuki$^{1}$, Moritaka Onitsuka$^{1}$, Yuki Asano$^{1}$, Kei Okada$^{1}$\\Koji Kawasaki$^{2}$, and Masayuki Inaba$^{1}$
  \thanks{$^{1}$ The authors are with the Department of Mechano-Informatics, Graduate School of Information Science and Technology, The University of Tokyo, 7-3-1 Hongo, Bunkyo-ku, Tokyo, 113-8656, Japan.
    {\texttt\small kawaharazuka@jsk.t.u-tokyo.ac.jp}
  }
  \thanks{$^{2}$ The author is associated with TOYOTA MOTOR CORPORATION.
  }
}
\begin{document}

\maketitle
\thispagestyle{empty}
\pagestyle{empty}

\begin{abstract}
  \switchlanguage%
  {%
    The flexible musculoskeletal hand is difficult to modelize, and its model can change constantly due to deterioration over time, irreproducibility of initialization, etc.
    Also, for object recognition, contact detection, and contact control using the hand, it is desirable not to use a neural network trained for each task, but to use only one integrated network.
    Therefore, we develop a method to acquire a sensor state equation of the musculoskeletal hand using a recurrent neural network with parametric bias.
    By using this network, the hand can realize recognition of the grasped object, contact simulation, detection, and control, and can cope with deterioration over time, irreproducibility of initialization, etc. by updating parametric bias.
    We apply this study to the hand of the musculoskeletal humanoid Musashi and show its effectiveness.
  }%
  {%
    柔軟な筋骨格ハンドはモデル化が難しく, また, 経年劣化やキャリブレーションのズレによって常にモデルが変化し得る.
    また, 物体認識や接触検知, 動的制御などをそれぞれのタスクに応じて学習されたネットワークを用いるのではなく, 一つのネットワークによって統合的に行えることが望ましい.
    そこで本研究では, パラメトリックバイアスを含む再帰的ニューラルネットワークを用いて, 筋骨格ハンドのセンサ値遷移に関する状態方程式を獲得する.
    これを用い, 把持物体認識, シミュレーション, 接触検知, 把持安定化制御を一つのネットワークで実現するとともに, パラメトリックバイアスを更新することで劣化やキャリブレーションのズレにも対応する.
    筋骨格ヒューマノイドMusashiのハンドに本研究を適用し, その有効性を示す.
  }%
\end{abstract}

\section{INTRODUCTION}\label{sec:introduction}
\switchlanguage%
{%
  Various robotic hands have been developed so far \cite{kochan2005shadowhand, kim2014roborayhand, deimel2016underactuatedhand, xu2016biohand, makino2018hand}.
  While there are many dexterous robotic hands with multiple rigid links \cite{kochan2005shadowhand, kim2014roborayhand}, flexible robotic hands \cite{deimel2016underactuatedhand, xu2016biohand, makino2018hand} are prevailing in terms of the adaptable grasping and impact response.
  The rubber hand structure of \cite{deimel2016underactuatedhand} is driven by air pressure, and the hand of \cite{xu2016biohand} is highly biomimetic with the elastic pulley system.
  In this study, we use the flexible and strong musculoskeletal hand \cite{makino2018hand} with machined spring fingers and a larger number of sensors than those hands.
  However, there are several problems concerning recognition of the grasped object, contact detection, and contact control with such a flexible hand \cite{deimel2016underactuatedhand, xu2016biohand, makino2018hand}.

  First, its modeling is difficult, and therefore, controls directly using geometric models are challenging.
  This problem is widely known now, and various learning-based object recognitions and contact controls have been studied.
  \cite{chebotar2016regrasping} predicted the success rate of grasping and realized regrasping using reinforcement learning.
  \cite{jain2019manipulation} trained imitation learning-based deep visuomotor policies and realized various manipulation tasks with a simulated five-fingered hand.
  While these studies are applied to only rigid hands, some studies can handle flexible hands as explained below.
  \cite{hoof2015reinforcement} applied reinforcement learning of in-hand manipulation with the under-actuated two-fingered robotic hand.
  \cite{homberg2019soft} realized the classification of the grasped objects using the pneumatic flexible robotic hand.
  However, these applications are limited, such as control on a two-dimensional plane or only grasping classification.

  Second, the modeling of the flexible hand is not only difficult but also can change constantly.
  Especially, soft materials such as rubber \cite{deimel2016underactuatedhand} and springs \cite{xu2016biohand, makino2018hand} significantly deteriorate over time, and also other materials such as wire and metal deteriorate due to poor use.
  Also, joint angle sensors sometimes cannot be attached to the flexible hand \cite{deimel2016underactuatedhand, xu2016biohand, makino2018hand}, and it is difficult to initialize the joint angles of the fingers.
  In the case of the hand \cite{makino2018hand} used in this study, there are no joint angle sensors, and so the origin of the joint angle is initialized when we manually extend the fingers.
  However, the initialization is hard to reproduce, and the irreproducibility of initialization changes the model we construct.
  In addition, when we modelize a sensor state transition of the hand by control commands, the model depends on the surrounding environment (e.g. grasped objects).
  While online learning methods of intersensory networks \cite{kawaharazuka2018online, kawaharazuka2019longtime} have been developed for musculoskeletal humanoids to solve these problems, online learning becomes difficult if the dimension to be learned increases, and so these methods can be applied to only static movements.
  Therefore, it is necessary to develop a method stably modifying and adapting the network to the current hand dynamics, including the difference of initialization, deterioration, and grasped objects, by changing only a small part of the network parameters.

  Third, as seen in the introduced studies, various components such as recognition of the grasped object, contact detection, and contact control have been developed individually \cite{chebotar2016regrasping, jain2019manipulation, hoof2015reinforcement, homberg2019soft}.
  While the individual network can specialize in each component, we consider that only one network representing sensor state transition is enough to handle these components in a relatively low layer.
  By using the integrated network, managing each network for each component is not necessary, and the parameter update making a certain component better can affect other components.
  Therefore, it is desirable to train only one integrated network for these components.
}%
{%
  これまで, 様々なロボットハンドが開発されてきた\cite{kochan2005shadowhand, kim2014roborayhand, deimel2016underactuatedhand, xu2016biohand, makino2018hand}.
  \cite{kochan2005shadowhand, kim2014roborayhand}のような多自由度でrigidなロボットハンドが多く存在する一方, \cite{deimel2016underactuatedhand, xu2016biohand, makino2018hand}にあるような柔軟ハンドが, 適応的な把持や衝撃吸収の観点から注目を浴びている.
  \cite{deimel2016underactuatedhand}はゴムで形成された構造が空気圧により駆動されており, \cite{xu2016biohand}はelastic pulley systemを使った, highly biomimeticなハンドである.
  本研究では, それらに比べて多数のセンサを持ち, 指自体が切削ばねにより構成される柔軟かつ力強い把持が可能な筋骨格柔軟ハンド\cite{makino2018hand}を用いる.
  しかし, これら柔軟なハンド\cite{deimel2016underactuatedhand, xu2016biohand, makino2018hand}は様々な利点の一方で, 把持による物体認識や接触検知, 動的制御等に現在もいくつかの問題点が存在する.

  まず, モデリングが難しいため, 幾何モデルを直接扱うような制御が難しいという点である.
  この問題点は現在は広く知られており, 学習型の把持物体認識や制御が盛んに研究されている.
  \cite{chebotar2016regrasping}は把持成功率の予測と強化学習による再把持を実現している.
  \cite{jain2019manipulation}はimitation learning based approach to train deep visuomotor policies for a variety of manipulation tasks with a simulated five fingered dexterous handしている.
  これらの研究はrigid linkなハンドで行われているが, 強化学習を用いていることもあり, 実機を長時間動かすまたは柔軟身体をシミュレーションするという課題がクリアできれば柔軟ハンドにも適用可能である.
  一方, 以下のように柔軟ハンドを使った学習型手法の例も存在する.
  \cite{hoof2015reinforcement}は劣駆動なロボットハンドに対して接触センサを用いて強化学習を適用し, 二次元平面上で二指によるIn-hand manipulationを実行している.
  \cite{homberg2019soft}は空気圧駆動の劣駆動柔軟ハンドによる把持物体の分類を行っている.
  しかし, 2次元平面上での制御であったり, 把持分類のみであったり, その適用先は限られている.

  次に, 柔軟ハンドはモデル化が難しいだけでなく, そのモデルが常に変わっていく点である.
  ゴムなどの柔らかい素材を用いている場合\cite{deimel2016underactuatedhand}は特に経年劣化が顕著に現れ, その他の素材の場合も劣悪な使用により劣化することが多い.
  また, 柔軟なハンドは関節角度センサ等を入れられないような場合が多く\cite{deimel2016underactuatedhand, xu2016biohand, makino2018hand}, ハンドの制御入力のキャリブレーションが難しい場合がある.
  本研究で用いる\cite{makino2018hand}のハンドも, 関節角度センサがないため, 人間が適当に指を伸ばした位置が0点となる.
  このような場合, キャリブレーションのズレによってモデルが変化してしまう.
  この問題に対して, 筋骨格ヒューマノイドではオンラインでセンサ間の写像を学習し続ける手法\cite{kawaharazuka2018online, kawaharazuka2018bodyimage, kawaharazuka2019longtime}が提案されているが, 学習する次元が多くなるとオンライン学習は困難となり, 静的な動作でのみ成功している.
  そのため, ネットワーク全体ではなく, その一部のみを変化させることでモデルを修正・適応させていくような方法が必要と考えられる.

  最後に, これまで述べたように柔軟ハンドの把持物体認識・接触検知・動的制御等, それらのコンポーネントはそれぞれ別々に設計されてきている\cite{chebotar2016regrasping, jain2019manipulation, hoof2015reinforcement, homberg2019soft}.
  それぞれindividualなnetworkを作る方法はそれぞれのtaskに特化することができるという利点がある一方, 本研究でtargetとなる比較的low layerなcomponentにはたった一つの状態遷移を表すnetworkさえあれば事足りると考えた.
  これにより, あるネットワークの更新が他のネットワークに影響を与えることができ, それぞれのネットワークを管理する必要もない.
  そのため, 一つのネットワークを学習し, そのネットワークのみを用いて様々なタスクができるようになることが望ましいと考える.
}%

\begin{figure}[t]
  \centering
  \includegraphics[width=0.8\columnwidth]{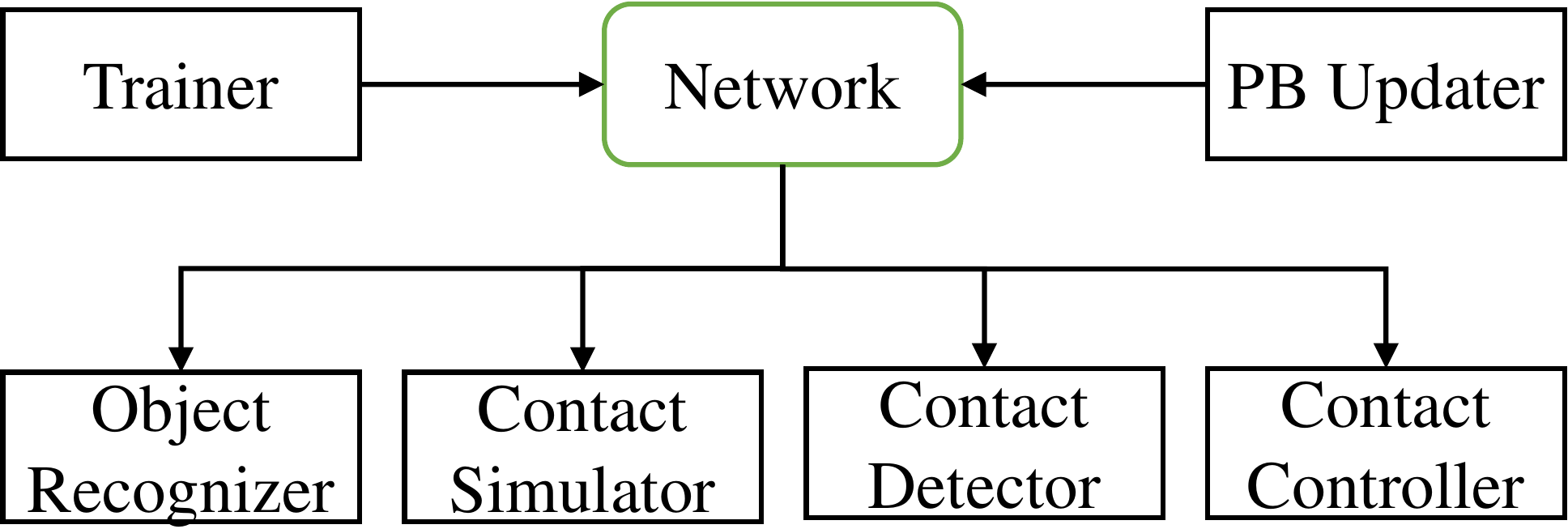}
  \caption{Overview of the developed system.}
  \label{figure:overview}
  \vspace{-3.0ex}
\end{figure}

\begin{figure*}[t]
  \centering
  \includegraphics[width=1.7\columnwidth]{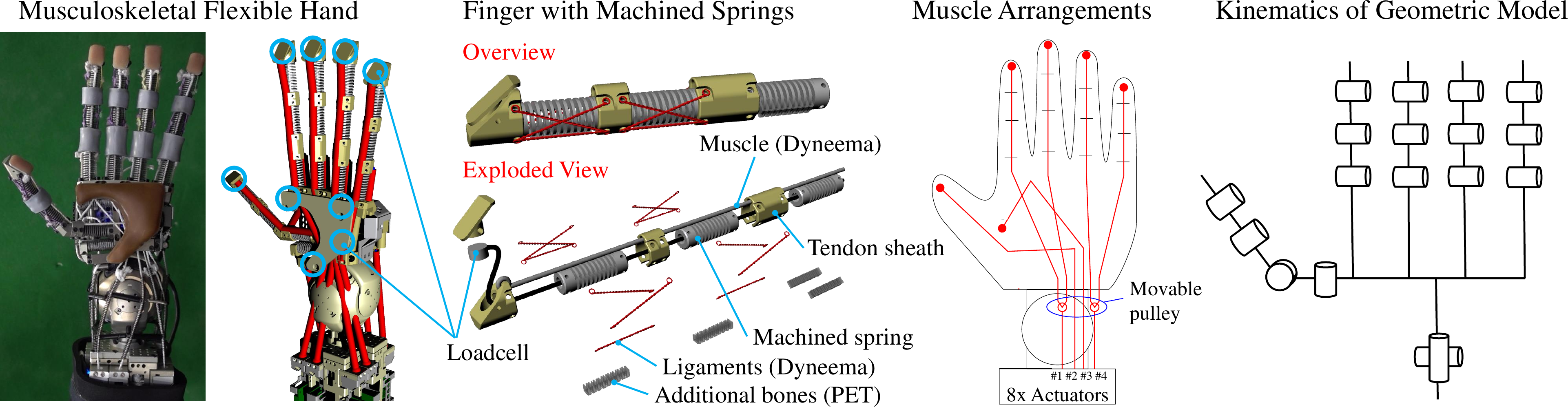}
  \caption{The five-finger flexible musculoskeletal hand \cite{makino2018hand} of the musculoskeletal humanoid Musashi \cite{kawaharazuka2019musashi}.}
  \label{figure:musculoskeletal-hand}
  \vspace{-3.0ex}
\end{figure*}

\switchlanguage%
{%
  In this study, we construct a sensor state equation represented by a recurrent neural network with parametric bias \cite{tani2002parametric} (Hand Dynamics Network, HADYNET), and use it for recognition of the grasped object, contact simulation, detection, and control (\figref{figure:overview}).
  We implicitly embed the information of deterioration, initialization, and grasped objects into the parametric bias (PB).
  By updating only PB, the network can adapt to the current hand dynamics and can cope with deterioration over time, irreproducibility of initialization, the change in grasped objects, etc.
  We apply this study to the flexible hand \cite{makino2018hand} of the musculoskeletal humanoid Musashi \cite{kawaharazuka2019musashi} and verify its effectiveness.
  The contributions of this study are shown below.
  \begin{itemize}
    \item Acquisition of a sensor state equation represented by a recurrent neural network with parametric bias for the flexible musculoskeletal hand.
    \item Coping with the change of the model due to deterioration over time, irreproducibility of initialization, the difference in grasped object, etc. by updating only parametric bias.
    \item Recognition of the grasped object, contact simulation, detection, and control using the trained network.
  \end{itemize}
}%
{%
  そこで本研究では, パラメトリックバイアスを含む再帰的ニューラルネットワーク\cite{tani2002parametric}によって表されたセンサ値の状態方程式(本研究では, Hand Dynamics Network, HADYNET)を構築し, これを用いて把持物体認識・シミュレーション・接触検知・把持安定化制御を総合的に行う(\figref{figure:overview}).
  パラメトリックバイアスのみを更新することで現在の状態に適応し, 経年劣化やキャリブレーション等にも対応する(\figref{figure:overview}のPB Updaterに対応する).
  本研究を筋骨格ヒューマノイドMusashi\cite{kawaharazuka2019musashi}の左手\cite{makino2018hand}に適用することで, その有効性を示す.
  本研究のコントリビューションは以下である.
  \begin{itemize}
    \item パラメトリックバイアスを含む再帰的ニューラルネットワークの学習によるハンドのセンサ状態方程式の獲得
    \item パラメトリックバイアスを用いた経年劣化やキャリブレーションのズレによるモデル変化への対応
    \item 本ネットワークを用いた把持物体認識・シミュレーション・接触検知・把持安定化制御手法の開発
  \end{itemize}
, etc}%

\section{Musculoskeletal Hand} \label{sec:musculoskeletal-hand}
\switchlanguage%
{%
  We show the musculoskeletal hand \cite{makino2018hand} used in this study, in \figref{figure:musculoskeletal-hand}.
  This hand attached to the musculoskeletal humanoid Musashi \cite{kawaharazuka2019musashi} has five fingers, and each finger joint is composed of a flexible machined spring.
  To make the machined springs anisotropic, Dyneema and PET plate imitating ligaments are attached to the side of the fingers.

  In the forearm of Musashi, there are eight muscle actuators \cite{kawaharazuka2017forearm}, and three and five of them are assigned to move the wrist and fingers, respectively.
  Two of the five tendons for fingers actuate the middle/index and ring/little fingers, respectively.
  Two tendons branched by a pulley control the two fingers at the same time.
  The other two of the five tendons actuate the thumb, and the last one can change the stiffness of fingers by pressing on the machined springs.

  Nine loadcells are distributed in each fingertip and the palm, and these arrangements are shown in the middle figure of \figref{figure:musculoskeletal-hand}.
  Muscle tension and length can be measured from the muscle actuator \cite{kawaharazuka2017forearm}.
  We represent the loadcell value as $\bm{f}_{contact}$, the current muscle length as $\bm{l}$, and muscle tension as $\bm{f}$.
  The kinematics of the geometric model of this hand is shown in the right figure of \figref{figure:musculoskeletal-hand}.
  The joints of machined springs are simply represented by rotational joints.
  Although finger joint angles cannot be measured, the wrist joint angles $\bm{\theta}$ can be measured by a joint module \cite{kawaharazuka2019musashi} in the wrist joint.

  The dimensions of $\bm{l}$ and $\bm{f}$ are 8, that of $\bm{f}_{contact}$ is 9, and that of $\bm{\theta}$ is 2.
  In the case of using only 4 muscle actuators \#1--\#4 directly involving finger joints, we represent these muscle tensions and lengths as $\bm{f}_{finger}$ and $\bm{l}_{finger}$.
}%
{%
  本研究で用いる筋骨格ハンド\cite{makino2018hand}を\figref{figure:musculoskeletal-hand}に示す.
  筋骨格ヒューマノイドMusashi \cite{kawaharazuka2019musashi}に搭載されたこのハンドは五指を有し, それぞれの指の関節は柔軟な切削バネによって構成されている.
  切削ばねは異方性を出すために靭帯を模したDyneemaとPETプレートが付属している.

  Musashi \cite{kawaharazuka2019musashi}の前腕には, 8本の筋アクチュエータ\cite{kawaharazuka2017forearm}が存在し, 手首に3本, 指に5本が割り当てられている.
  指の5本の腱のうち2本はそれぞれ人差し指と中指, 薬指と小指を駆動しており, 2つの指をプーリで分岐することで制御している.
  また, うち2本は親指を駆動している.
  最後の一本は張力を高めることで切削ばねを押しつぶし, 指の剛性を変化させることができる.

  それぞれの指の先端, 手のひらには9つの接触センサとしてのロードセルが分布しており, その配置は\figref{figure:musculoskeletal-hand}の中図のようになっている.
  また, 筋アクチュエータ\cite{kawaharazuka2017forearm}からは筋長, 筋張力を測定することができる.
  このロードセル値を$\bm{f}_{contact}$, エンコーダから得られる現在筋長を$\bm{l}$, 筋張力値を$\bm{f}$とする.
  加えて, 手首には関節モジュール\cite{kawaharazuka2019musashi}があり, 指の関節角度は測れないものの, 手首の関節角度$\bm{\theta}$は測定することができる.

  $\bm{l}, \bm{f}$は8次元, $\bm{f}_{contact}$は9次元, $\bm{\theta}$は2次元である.
  また, 一部で指に直接関係のある4つの筋\#1--\#4のみを用いる場合は, これを$\bm{f}_{finger}, \bm{l}_{finger}$と表し, それぞれ4次元とする.
}%

\section{Hand Dynamics Network} \label{sec:proposed}
\begin{figure}[t]
  \centering
  \includegraphics[width=0.8\columnwidth]{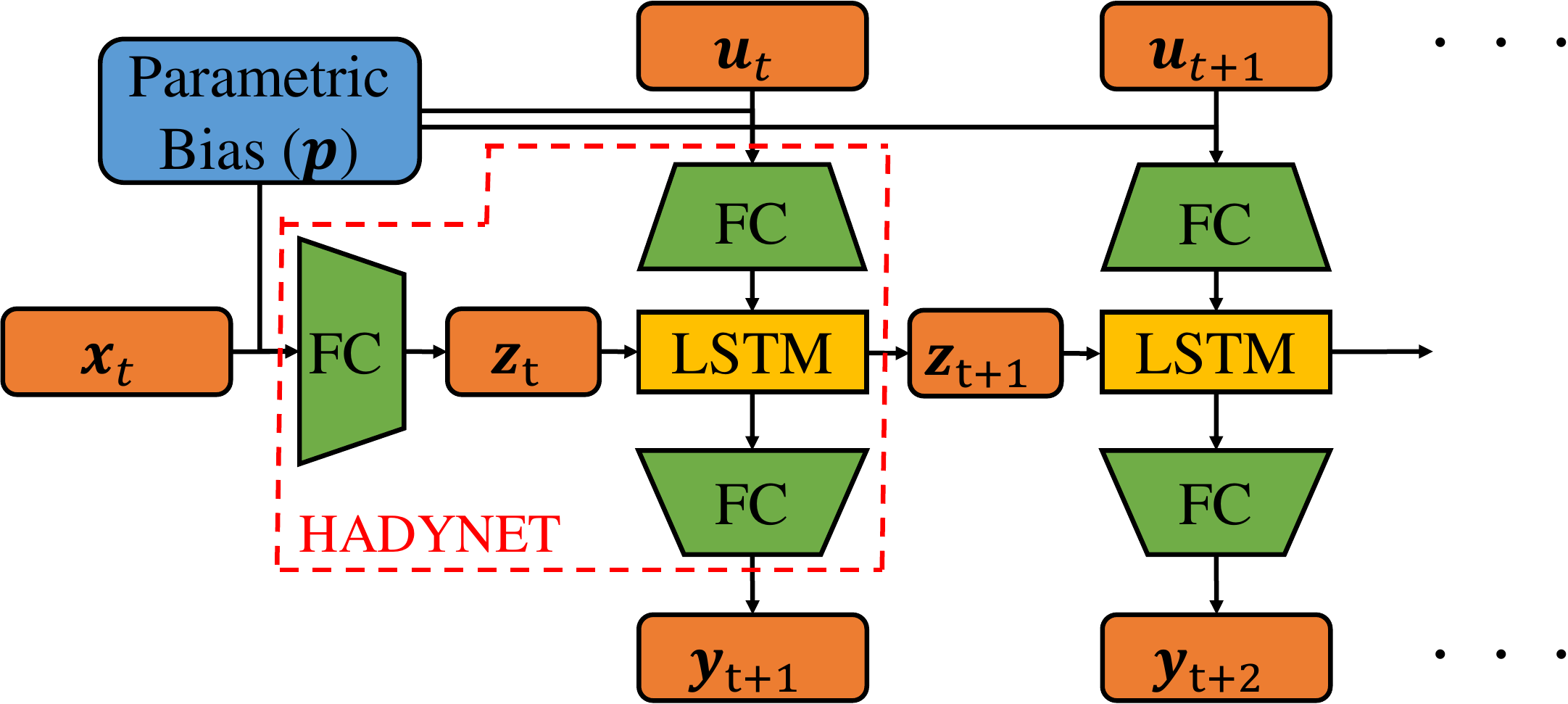}
  \caption{Network structure of hand dynamics network (HADYNET).}
  \label{figure:network-structure}
  \vspace{-3.0ex}
\end{figure}

\begin{figure*}[t]
  \centering
  \includegraphics[width=1.8\columnwidth]{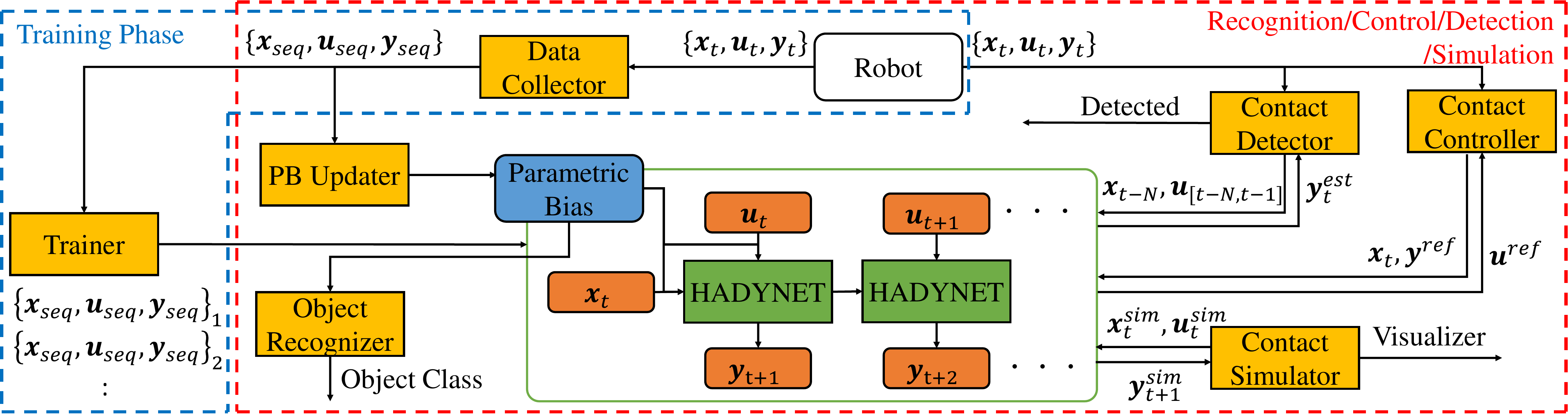}
  \caption{The overall system using the hand dynamics network.}
  \label{figure:system}
  \vspace{-3.0ex}
\end{figure*}

\switchlanguage%
{%
  We show the detailed network structure in \figref{figure:network-structure} and the whole system in \figref{figure:system}.
}%
{%
  本研究では, Hand Dynamics Network (HADYNET)のネットワーク構造, その学習, パラメトリックバイアスの更新, これを用いた把持物体認識について説明する.
  その後, HADYNETを用いた接触シミュレーション, 接触検知, 接触制御について述べていく.
  全体のシステム図を\figref{figure:system}に示す.
}%

\subsection{Network Structure} \label{subsec:network-structure}
\switchlanguage%
{%
  HADYNET is the network making use of the recurrent neural network with parametric bias (RNNPB) proposed by J. Tani \cite{tani2002parametric}.
  Several studies using RNNPB have been conducted \cite{tani2004parametric, ogata2005extracting} so far, and RNNPB has been used to embed multiple dynamics of various motion sequences into one network.
  In this study, we make use of RNNPB as the sensor state equation with multiple dynamics caused by deterioration over time, irreproducibility of initialization, the difference in grasped objects, etc.

  HADYNET is represented by functions $\bm{h}_{\{1, 2\}}$ as below,
  \begin{align}
    \bm{z}_{t} &= \bm{h}_{1}(\bm{x}_{t}, \bm{p}) \label{eq:hdnet-1}\\
    \bm{y}_{t+1} &= \bm{h}_{2}(\bm{z}_{t}, \bm{u}_{t}, \bm{p}) \label{eq:hdnet-2}
  \end{align}
  where $\bm{x}$ is the initial sensor state of the hand, $\bm{z}$ is the hidden state, $\bm{u}$ is the control command, $\bm{y}$ is the observed sensor state, $\bm{p}$ is the parametric bias (PB), $t$ is the current time step, and $\bullet_{t}$ is the value at the time step $t$.
  In this study, $\bm{x}$ represents $(\bm{l}^{T}, \Delta{\bm{l}}^{T}, \bm{f}^{T}, \bm{f}^{T}_{contact}, \bm{\theta}^{T}, \Delta{\bm{\theta}}^{T})^{T}$ ($\Delta{\{\bm{l}, \bm{\theta}\}}$ is the difference value from the previous time step, and the dimension of $\bm{x}$ is 37).
  Also, $\bm{u}$ represents $\Delta\bm{l}^{ref}$ ($\Delta\bm{l}^{ref}$ is the change of the target muscle length, and the dimension of $\Delta\bm{l}^{ref}$ is 8), $\bm{y}$ represents $(\bm{l}^{T}, \bm{f}^{T}, \bm{f}^{T}_{contact}, \bm{\theta}^{T})^{T}$ (the dimension of $\bm{y}$ is 27).
  We call \equref{eq:hdnet-1} and \equref{eq:hdnet-2} HADYNET.
  Thus, HADYNET represents the state equation of all the sensors equipped with the hand by the control command of target muscle velocity.

  The detailed network structure is shown in \figref{figure:network-structure}.
  FC represents a fully-connected layer, and LSTM represents Long Short-Term Memory \cite{hochreiter1997lstm}.
  The number of units of LSTM is 128, and the dimension of PB is 8.
  The number of units of FC can be determined from the input and output dimensions, and after feeding the input value into FC, the value is fed into Batch Normalization \cite{ioffe2015batchnorm} and ReLU \cite{nair2010relu}.
  The cell state of LSTM is initialized to 0.

  In this study, the interval of the adjacent time steps is 0.2 seconds.
}%
{%
  HADYNETは, J. Taniが考案したパラメトリックバイアスを含む再帰的ニューラルネットワーク(RNNPB)\cite{tani2002parametric}を利用している.
  これまでRNNPBを用いたいくつかの研究がなされてきたが\cite{tani2004parametric, ogata2005extracting}, 複数のattractorを有する様々な動作シーケンスを一つのネットワークに落としこむために持ち入られてきた.
  本研究ではこれを, 把持物体やキャリブレーション, 経年劣化等の様々な状態におけるattractorを包含する状態方程式として利用している.

  HADYNETを簡易的に関数$\bm{h}_{\{1, 2\}}$で表すと以下のようになる.
  \begin{align}
    \bm{z}_{t} = \bm{h}_{1}(\bm{x}_{t}, \bm{p}) \label{eq:hdnet-1}\\
    \bm{y}_{t+1} = \bm{h}_{2}(\bm{z}_{t}, \bm{u}_{t}, \bm{p}) \label{eq:hdnet-2}
  \end{align}
  ここで$\bm{x}$はハンドの初期状態, $\bm{z}$は隠れ状態, $\bm{u}$は制御入力, $\bm{y}$は観測状態, $\bm{p}$はパラメトリックバイアス, $\bm{t}$は現在のタイムステップを指す.
  本研究において, $\bm{x}$は$(\bm{l}^{T}, \Delta{\bm{l}}^{T}, \bm{f}^{T}, \bm{f}^{T}_{contact}, \bm{\theta}^{T}, \Delta{\bm{\theta}}^{T})^{T}$を表す($\Delta{\{\bm{l}, \bm{\theta}\}}$は前ステップからの差分, つまり速度を表し, $\bm{x}$は本研究では37次元である).
  また, $\bm{u}$は$\Delta\bm{l}^{ref}$ ($\bm{l}^{ref}$は指令筋長を表し, $\Delta\bm{l}^{ref}$は本研究では8次元である), $\bm{y}$は$(\bm{l}^{T}, \bm{f}^{T}, \bm{f}^{T}_{contact}, \bm{\theta}^{T})$ ($\bm{y}$は本研究では27次元である)を表す.
  この\equref{eq:hdnet-1}, \equref{eq:hdnet-2}を合わせたものを, 本研究ではHADYNETと呼ぶこととする.
  つまり, HADYNETはハンドに備わる全センサの筋長速度指令に関する状態方程式を表す.
  後に説明するが, ネットワークの重みを$\bm{W}$とすると, 本研究におけるtrainingは$\bm{W}$と$\bm{p}$を, 把持物体認識は$\bm{p}$を, 接触制御は$\bm{u}$を, 与えられた損失関数から誤差逆伝播法によって更新することに相当している.

  具体的なネットワーク構造を\figref{figure:network-structure}に示す.
  ここで, FCはFully Connected Layer, LSTMはLong Short-Term Memory \cite{hochreiter1997lstm}を表す.
  LSTMのユニット数は128, パラメトリックバイアスは8次元とする.
  FCはそれぞれ入力と出力の値から次元数は一意に決定されており, Fully Connected Layerを通した後に, Batch Normalization \cite{ioffe2015batchnorm}, ReLU \cite{nair2010relu}をかけている.
  また, 特に記載がない場合は, LSTMにおける初期のcell stateは0で初期化する.

  本研究ではタイムステップごとの間隔を0.2秒としている.
}%

\subsection{Training of HADYNET} \label{subsec:training}
\switchlanguage%
{%
  To embed multiple dynamics into PB, we collect various data with various dynamics, and train HADYNET as each data corresponds to different PB.

  First, we prepare the hand dynamics state $k$ with different initializations, deteriorations, and grasped objects.
  We conduct various motions with the hand of each dynamics state $k$, and obtain the $N$ steps motion data of $D_{k} = \{\bm{x}_{[0, N)}, \bm{u}_{[0, N)}, \bm{y}_{[0, N)}\}_{k}$.
  In \figref{figure:system}, $\bullet_{[0, N)}$ is abbreviated to $\bullet_{seq}$.

  Second, we convert the collected data.
  We determine $N^{train}$, how many time steps are predicted by HADYNET.
  We extract $N^{train}$ steps of consecutive data from $D_{k}$ every one time step, and we represent all the extracted data as $D'_{k}=\{\{\bm{x}_{0}, \bm{u}_{[0, N^{train})}$, $\bm{y}_{[1, N^{train}+1)}\}_{k}, \{\bm{x}_{1}, \bm{u}_{[1, N^{train}+1)}, \bm{y}_{[2, N^{train}+2)}\}_{k}, \cdots\}$.
  Regarding $\bm{x}$, only the initial $\bm{x}$ is necessary.
  We execute this process for all the collected data and prepare $D'_{k=\{1, 2, ..., K\}}$ ($K$ represents the number of the collected hand dynamics state).

  Third, we train HADYNET using $D'$.
  We shuffle $D'$ and train HADYNET by setting the batch size $N^{train}_{batch}$ and the epoch $N^{train}_{epoch}$.
  We use $\bm{p}_{k}$ as PB of the hand dynamics state $k$ (PB is the same among the same hand dynamics state but different among the different hand dynamics state), and this $\bm{p}_{k}$ is implicitly trained with HADYNET.
  All $\bm{p}_{k}$ are initialized to $\bm{0}$ before training.

  In this study, $N$ is different regarding each hand dynamics state, and we set $N^{train}=10$, $N^{train}_{batch}=100$, and $N^{train}_{epoch}=200$.
}%
{%
  訓練用データを蓄積し, HADYNETを学習させる.
  本研究では把持物体やキャリブレーション, 経年劣化等の変化の様々な状態におけるattractorを包含する状態方程式を得るために, それぞれの状態を別のタスクとしてネットワークに学習させる.

  まず, キャリブレーション・把持物体を変化させた, ある一つのハンド状態を用意し, これをタスク$k$とする.
  様々な動作を行い, その際の$N$ステップ分のデータ$D_{k} = \{\bm{x}_{[0, N)}, \bm{u}_{[0, N)}, \bm{y}_{[0, N)}, k\}$を取得する.
  本研究では, 2種類のランダムな動作(Random-1, Random-2)を行う.
  1つめ(Random-1)は, 指令関節角度に対応する指令筋長が得られるような幾何モデルを用意し, 設定したランダムな指令関節角度を筋長に変換し, ランダムな秒数をかけて動かすことを繰り返す動作である.
  2つめ(Random-2)は, 指を曲げて物体を把持し, その状態で指に直接関係する$\bm{l}_{finger}$だけランダムに変化させる動作である(もし何も把持しないタスクの場合は適当な量指を曲げる).
  後者は後に説明する\secref{subsec:control}において重要となるデータであり, 握り方の違いによる接触の変化を得ることができる.

  次に, HADYNETで何ステップ先まで予測をさせるかを決める$N_{train}$を決定する.
  そして, $D_{k}$から1ステップごとに連続する$N_{train}$ステップ分のデータを取り出し, このデータをを$D'_{k}=\{\{\bm{x}_{0}, \bm{u}_{[0, N_{train})}$, $\bm{y}_{[1, N_{train}+1)}, k\}, \{\bm{x}_{1}, \bm{u}_{[1, N_{train}+1)}, \bm{y}_{[2, N_{train}+2)}, k\}, \cdots\}$とする.
  なお, $\bm{x}$は最初のステップのみ必要なため, はじめのタイムステップだけ取り出している.
  この試行を様々なハンド状態について行い, $D'_{k=\{1, 2, ..., K\}}$用意する($K$は行ったタスクの数を表す).

  最後に, このデータを用いてHADYNETを学習させる.
  得られたデータを全てシャッフルし, バッチ数を$C_{batch, train}$, エポック数を$C_{epoch, train}$として学習させる.
  このとき, パラメトリックバイアス$\bm{p}$の値は$\bm{p}_{k}$を用い, タスクごとに共通であるがタスク間では異なるパラメータとし, これをHADYNETと同時にimplicitに学習させる.

  本研究では, $N$はタスクごとに異なり, $N_{train}=10$, $C_{batch, train}=100$, $C_{epoch, train}=200$として学習させている.
}%

\subsection{Update of Parametric Bias} \label{subsec:update}
\switchlanguage%
{%
  Although the network weight of HADYNET and PB are trained in \secref{subsec:training}, we do not need the trained PB when using HADYNET except for in the object recognition task.
  When changing the grasped objects or initializing again, we collect the data $D'$ regarding the current hand dynamics state, fix the network weight of HADYNET, and update only PB.
  As explained in \secref{sec:introduction}, by updating only PB rather than all the network weight, we can adapt HADYNET to be close to the current hand dynamics state, while avoiding an over-fitting problem.
  Because only the low dimensional space of PB is updated, only the hand dynamics of the network is adapted while keeping the overall sensor state equation.
  In \secref{subsec:training}, if we train HADYNET with various grasped objects, initializations, and deterioration, the multiple dynamics states are constructed in PB.
  By searching the PB fitting to the current sensor transition, we can obtain the current hand dynamics state.
  We represent the batch size as $N^{update}_{batch}$ and the number of epochs as $N^{update}_{epoch}$, and use MomentumSGD as the update rule.

  Also, although the method to train HADYNET offline using the collected $D'$ is more stable, the online update of PB is enabled because the over-fitting problem can be avoided.
  We determine the data size $N^{online}_{thre}$ that starts the online update of PB, and the maximum accumulated data size $N^{online}_{max}$, by which the update can be executed online.
  We accumulate data as in \secref{subsec:training}, and when the accumulated data size exceeds $N^{online}_{thre}$, PB starts to be updated by setting the batch size $N^{online}_{batch}$ and the number of epochs $N^{online}_{epoch}$.
  When the accumulated data size exceeds $N^{online}_{max}$, the oldest data is removed.

  In this study, $N^{\{update, online\}}_{batch}=N$ ($N$ is the number of data in $D'$), $N^{update}_{epoch}=20$, $N^{online}_{epoch}=3$, $N^{online}_{thre}=100$, and $N^{online}_{max}=200$.
}%
{%
  HADYNET, Parametric Biasが学習されたが, 実際にHADYNETを使用するときは, \secref{subsec:training}で学習されたParametric Biasは破棄する.
  改めてハンドに道具を持たせたり, キャリブレーションをしたときに, もう一度$D'$のデータを取得し(この状態におけるデータのみで良い), HADYNETは固定して, Parametric Biasのみ学習させる.
  \secref{sec:introduction}でも説明したように, 全体ではなく, Parametric Biasという一部を学習させることによって, 過学習せずに適切かつ素早くネットワークを現在の状態に適合させることができる.
  \secref{subsec:training}において, 様々な物体把持, キャリブレーション, 経年劣化等の状態について学習をしておけば, それらの状態がParametric Biasに構築される.
  そのParametric Bias内部を現在のデータから探索することで, 現在のdynamicsの状態を得ることができるのである.
  このときのバッチサイズを$C_{batch, update}$, エポック数を$C_{epoch, update}$とし, 更新則はMomentumSGDを用いる.

  また, 上記のように一度$D'$を取得してから学習させるのが最も安定しているが, 過学習を避けられるため, オンラインでの更新も可能である.
  オンライン学習を始めるデータ数$C_{thre, online}$と, オンラインで学習が完了する程度の蓄積する最大データ数$C_{max, online}$を決める.
  データを上記と同じように貯めていき, $C_{thre, online}$を超えたらバッチサイズを$C_{batch, online}$, エポック数を$C_{epoch, online}$として学習し始める.
  データが$C_{max, online}$を超えたら, 一番古いデータを削除することを繰り返す.

  本研究では, $C_{batch, \{update, online\}}=N$ ($N$は$D'$のデータ数とする), $C_{epoch, update}=20$, $C_{epoch, online}=3$, $C_{thre, online}=100$, $C_{max, online}=200$とする.
}%

\subsection{Recognition of Grasped Object} \label{subsec:recognition}
\switchlanguage%
{%
  The robot can recognize the grasped object by making use of the PB updated in \secref{subsec:update}.
  Regarding the $K$ hand dynamics states used in \secref{subsec:training}, we save the trained PB $\bm{p}_{k}$ and the name of each grasped object.
  We can visually plot these $\bm{p}_{k}$ by reducing the dimension of PB to two-dimensional space using principal component analysis (PCA).
  After updating PB in \secref{subsec:update}, we convert the updated $\bm{p}$ by the converter of PCA as above and plot it in the two-dimensional space.
  We can visually obtain the current object name from the position of $\bm{p}$ in $\bm{p}_{k}$.
  Also, by the nearest neighbor method, the robot can recognize which object is the closest to the current grasped object.
  This object recognition is possible because the difference of hand dynamics caused by initialization and deterioration is smaller than that caused by the difference in grasped objects.
  This was found through subsequent experiments, as the difference of grasped objects appeared like in \figref{figure:object-recognition} and \figref{figure:online-pb} even if we obtain PB while changing the initialization and grasped objects.
}%
{%
  \secref{subsec:update}において学習されたParametric Biasを用いることで, 把持物体の認識を行うことができる.
  \secref{subsec:training}において用いた$K$個のタスクに対して, 学習されたパラメトリックバイアス$p_{k}$と, それぞれのタスクにおいて把持している物体の名前を保存しておく.
  $\bm{p}_{k}$の値を主成分分析により2次元に落とし, 2次元平面にプロットしておく.
  \secref{subsec:update}のおいて学習された$\bm{p}$に先のPCAで得られた変換をかませ, 同様に2次元平面にプロットする.
  これにより, 現在の$\bm{p}$の位置から, 視覚的に現在把持している物体がどの物体に近いかを得ることができる.
  また, 最近傍法により, 最も$\bm{p}$に近い$\bm{p}_{k}$を現在把持している物体として, 認識することが可能である.
}%

\subsection{Contact Simulation} \label{subsec:simulation}
\switchlanguage%
{%
  The contact simulation is executed only by forwarding HADYNET.
  We set the initial sensor state $x^{sim}_{t}$, and by feeding the control command $u^{sim}_{t}$ into HADYNET, $y^{sim}_{t+1}$ can be obtained.
  We can visually simulate the force and position of the hand by drawing $\bm{f}$ as color gradation, $\bm{f}_{contact}$ as the size of the force arrows, and $\bm{\theta}$ as the joint angle.
  Also, just by changing PB, we can observe the change of hand dynamics with various grasped objects and initializations.
  This simulation can be used for reinforcement learning.
}%
{%
  接触のシミュレーションはHADYNETの順伝播のみで行うことができる.
  初期状態$x^{sim}_{t}$を設定し, コマンド$u^{sim}_{t}$を送ることで, $y^{sim}_{t+1}$を得ることができる.
  $\bm{f}$は筋の色で, $\bm{f}_{contact}$は力の矢印の大きさで, $\bm{\theta}$は関節角度で描画することで, ハンドの動作指令における力や位置のシミュレーションを行うことができる.
  また, Parametric Biasを変更することで, 物体の有無や種類によるハンドのdynamicsの変化を観察し, 強化学習等にも用いることができると考える.
}%

\subsection{Contact Detection} \label{subsec:detection}
\switchlanguage%
{%
  From the prediction error of HADYNET, we can execute contact detection (anomaly detection) of the hand.
  First, before contact detection, we collect $D'$ and update PB as in \secref{subsec:update} without grasping anything.
  We determine the time steps $N^{detect}$ ($N^{detect} \leq N^{train}$) to be predicted by HADYNET, and regarding $D'$, we calculate the average and covariance matrix of the prediction error of HADYNET after $N^{detect}$ time steps.
  In detail, we extract $\bm{x}_{i-N^{detect}}$ and $\bm{u}_{[i-N^{detect}, i)}$ when setting $i$ as the last time step of the $N^{detect}$ steps sequence, calculate the error between $\bm{y}^{est}_{i}$ predicted by using the extracted data with HADYNET and $\bm{y}_{i}$ in $D'$, and obtain the average $\mu_{e}$ and the covariance matrix $\Sigma_{e}$ of the errors regarding all the sequence in $D'$.

  Second, we will explain the actual contact detection phase.
  We always accumulate $\bm{x}$ and $\bm{u}$ from the current time step $t$ to $t-N^{detect}$.
  At $t$, we extract $\bm{x}_{t-N^{detect}}$ and $\bm{u}_{[t-N^{detect}, t)}$, and predict $\bm{y}^{est}_{t}$ through HADYNET.
  Then, we obtain $\bm{y}_{t}$ and calculate Mahalanobis distance as shown below,
  \begin{align}
    d = \sqrt{(\bm{e}_{t}-\mu_{e})^{T}\Sigma^{-1}_{e}(\bm{e}_{t}-\mu_{e})}
  \end{align}
  where $\bm{e}_{t}=\bm{y}^{est}_{t}-\bm{y}_{t}$.
  When $d$ exceeds the threshold $C^{detect}$, we regard that a motion different from the prediction occurs, and so the unpredicted contact is detected.
  Although we can use 3$\sigma$ of $d$ in $D'$ as $C^{detect}$, as too many contacts should not be detected, we set a higher constant value.

  In this study, we set $C^{detect}=100$.
}%
{%
  HADYNETにおける予測誤差から, ハンドにおける接触検知(異常検知)を行うことができる.
  まず, 接触検知を行う前に\secref{subsec:update}を行う.
  そのあと, HADYNETで予測するタイムステップ数$N_{detect}$ ($N_{detect} \leq N_{train}$)を決め, データ$D'$に対して, $N_{detect}$ステップ後における予測誤差の平均と共分散行列を計算しておく.
  つまり, ある一つデータの終点を$i$として, $\bm{x}_{i-N_{detect}}$と$\bm{u}_{[i-N_{detect}, i)}$を取り出し, それを用いてHADYNETにより予測された$\bm{y}^{est}_{i}$とデータセット内の$\bm{y}_{i}$の誤差を計算して, 全データについてこれの平均$\mu$と共分散行列$\Sigma$を計算するということである.

  次に, 実際の接触検知について説明する.
  現在のタイムステップ$t$から$N_{detect}$前までの$\bm{x}$と$\bm{u}$を常に保存しておく.
  ここで, 現在$t$において, $\bm{x}_{t-N_{detect}}$と$\bm{u}_{[t-N_{detect}, t)}$を取り出し, HADYNETにより, $\bm{y}^{est}_{t}$を予測する.
  ここで, $\bm{y}_{t}$を得て, 以下のマハラノビス距離を計算する.
  \begin{align}
    d = \sqrt{(\bm{y}_{t}-\mu)^{T}\Sigma(\bm{y}^{est}_{t}-\mu)}
  \end{align}
  この$d$が閾値$C_{detect}$を超えたとき, 予想とは違う動きが発生, つまり接触が検知されたと見なす.
  この$C_{detect}$はデータ$D'$における$d$の分散の3$\sigma$等を用いることも可能だが, 余計な接触が検知されないよう, 本研究では高めの値を設定している.

  本研究では$C_{detect}=100$とする.
}%

\subsection{Contact Control} \label{subsec:control}
\switchlanguage%
{%
  By optimizing the control command to make the predicted sensor state close to the target value, the robot can realize the target contact state.
  Here, the contact state means the values of $\bm{f}$ and $\bm{f}_{contact}$.
  This control is a method applying \cite{kawaharazuka2020regrasp} to HADYNET.

  First, we determine the number of time steps $N^{control}$ ($N^{control} \leq N^{train}$) to be predicted (control horizon) and the target contact state $\bm{y}^{ref}$.
  Second, we determine the control command $\bm{u}^{opt}_{[t, t+N^{control})}$ before optimization (we represent it as $\bm{u}^{opt}_{seq}$ below).
  By obtaining $\bm{x}_{t}$ at the current time step and feeding it with $\bm{u}^{opt}_{seq}$ into HADYNET, the predicted sensor state of $\bm{y}^{est}_{[t+1, t+N^{control}+1)}$ is obtained.
  Then, the loss of $L$ is calculated by the loss function $h_{loss}$, and $\bm{u}^{opt}_{seq}$ is optimized as shown below,
  \begin{align}
    L &= h_{loss}(\bm{y}^{est}_{seq}, \bm{y}^{ref}_{seq})\\
    \bm{g} &= \partial{L}/\partial{\bm{u}^{opt}_{seq}}\\
    \bm{u}^{opt}_{seq} &\gets \bm{u}^{opt}_{seq} - \alpha\bm{g}/||\bm{g}||_{2}
  \end{align}
  where $\bm{y}^{est}_{seq}$ represents $\bm{y}^{est}_{[t+1, t+N^{control}+1)}$, $\bm{y}^{ref}_{seq}$ represents a vector vertically arranging $N^{control}$ vectors of $\bm{y}^{ref}$, $||\bullet||_{2}$ represents L2 norm, and $\alpha$ is an update rate.
  $\bm{u}^{opt}_{seq}$ is updated using $N^{control}_{batch}$ kinds of $\alpha$, $\bm{u}^{opt}_{seq}$ with the minimum $L$ in the batch is adopted, and the gradient $g$ is calculated again.
  This is repeated $N^{control}_{epoch}$ times.
  Also, we use $\bm{u}^{opt}_{\{t, \cdots, t+N^{control}-1, t+N^{control}-1\}}$, in which $\bm{u}^{opt}_{[t-1, t+N^{control}-1)}$ optimized at the previous time step is shifted and its last term is replicated, as $\bm{u}^{seq}_{control}$.
  After optimization, $\bm{u}^{opt}_{t}$ is sent to the actual robot.

  Here, we consider the design of the loss function $h_{loss}$.
  In this study, we mainly consider the stabilization of the grasp and aim to keep the initial contact state at all times when grasping the tool object.
  Therefore, $\bm{y}^{ref}$ is the initial value $\bm{y}^{init}$ when grasping the tool.
  For example, although the contact state and the grasp condition gradually change when grasping a hammer and hitting with it, we aim to inhibit the change.
  In this case, we design the loss function as shown below,
  \begin{align}
    &h_{loss}(\bm{y}^{est}_{seq}, \bm{y}^{ref}_{seq}) = h_{loss, 1}(\bm{y}^{est}_{seq}, \bm{y}^{ref}_{seq}) + h_{loss, 2}(\bm{y}^{est}_{seq}, \bm{y}^{ref}_{seq})\\
    &h_{loss, 1}(\bm{y}^{est}_{seq}, \bm{y}^{ref}_{seq}) = ||\bm{w}^{T}_{1}(\bm{F}^{est}_{seq}-\bm{F}^{ref}_{seq})||^{2}_{2}\\
    &\bm{w}_{1}[i] = \begin{cases}
      1.0\;\;\;(\bm{F}^{est}_{seq}[i] \geq \bm{F}^{ref}_{seq}[i])\nonumber\\
      \beta\;\;\;\;\;\;(otherwise)\nonumber
    \end{cases}\\
    &h_{loss, 2}(\bm{y}^{est}_{seq}, \bm{y}^{ref}_{seq}) = w_{2}||\bm{\theta}^{est}_{seq}-\bm{\theta}^{ref}_{seq}||^{2}_{2} + w_{3}||\bm{l}^{est}_{seq}-\bm{l}^{ref}_{seq}||^{2}_{2}
  \end{align}
  where, $\bm{F}^{\{est, ref\}}_{seq}$ represents a vector vertically arranging $\bm{f}$ and $\bm{f}_{contact}$ extracted from $\bm{y}^{\{est, ref\}}_{seq}$, and $\{\bm{\theta}, \bm{l}\}^{\{est, ref\}}_{seq}$ represents a vector vertically arranging $\{\bm{\theta}, \bm{l}\}$ extracted from $\bm{y}$.
  Also, $w_{\{2, 3\}}$ is a constant weight, $\bm{w}_{1}$ is a weight vector, $\bm{w}_{1}[i]$ is the $i^{\textrm{th}}$ element of $\bm{w}_{1}$, and $\beta$ is a constant weight.
  The design of $w_{1}$ considers the characteristics of the contact sensors.
  Although the sensor values of the contact and muscle tension sensors change just until 0 in the minus direction, the values can largely change until the rated values in the plus direction.
  When setting $\beta=1.0$, as the result of optimization, the contact state tends to change in the minus direction and the initial contact state cannot be kept.
  Therefore, by setting $\beta>1.0$, we generate the control command to keep the initial contact state.
  At the same time, by limiting joint angles and muscle lengths by $h_{loss, 2}$, we can obtain the control command which does not largely change the muscle length and joint angle while keeping the contact state.

  In this study, we set $N^{control}=8$, $N^{control}_{batch}=4$, $N^{control}_{epoch}=3$, $\beta=3.0$, $w_{2}=1.0$, and $w_{3}=1.0$.
}%
{%
  HADYNETによる予測を指令値に近づけるように制御指令を最適化することで, 指令した接触状態を実現することができる.
  これは, \cite{kawaharazuka2020regrasp}で提案された接触制御を本研究のネットワークに適用したものである.

  まず, HADYNETで予測するタイムステップ$N_{control}$ ($N_{control} \leq N_{train}$), 接触状態の指令値$\bm{y}^{ref}$を決める.
  次に, 最適化前の制御指令値$\bm{u}^{opt}_{[t, t+N_{control})}$ (以降では$\bm{u}^{opt}_{seq}$とする)を決定する.
  現在のタイムステップ$t$における$\bm{x}_{t}$を取得し, $\bm{u}^{opt}_{seq}$とともにHADYNETに入力することで, 予測値$\bm{y}^{est}_{[t+1, t+N_{control}+1)}$を得る.
  そして, 損失関数$h_{loss}$により損失$L$を計算し, 以下のように$\bm{u}^{opt}$を最適化していく.
  \begin{align}
    L &= h_{loss}(\bm{y}^{est}_{seq}, \bm{y}^{ref}_{seq})\\
    \bm{g} &= \partial{L}/\partial{\bm{u}^{opt}_{seq}}\\
    \bm{u}^{opt}_{seq} &\gets \bm{u}^{opt}_{seq} - \alpha\bm{g}/||\bm{g}||_{2}
  \end{align}
  ここで, $\bm{y}^{est}_{seq}$は$\bm{y}^{est}_{[t+1, t+N_{control}+1)}$を, $\bm{y}^{ref}_{seq}$は$\bm{y}^{ref}$を縦に$N_{control}$個並べたベクトル, $\alpha$は更新率を表す.
  実際には, $C_{batch, control}$種類の$\alpha$を使って$\bm{u}^{opt}_{seq}$を更新し, そのバッチの中で最も$L$が小さかったものを採用し, また勾配を計算することを$C_{epoch, control}$回繰り返す.
  また, $\bm{u}^{seq}_{control}$は前ステップで最適化された$\bm{u}^{opt}_{[t-1, t+N_{control}-1)}$を一つ左へシフトし, 最終項を複製した$\bm{u}^{opt}_{\{t, \cdots, t+N_{control}-1, t+N_{control}-1\}}$を用いる.
  最適化後, $\bm{u}^{seq}_{t}$が実機に送られる.

  ここで, 損失関数$h_{loss}$の設計について考える.
  本研究では, 主に把持安定化について考え, 道具を把持したときの初期の接触を, 常に保ち続けるような制御を行うことを目的とする.
  つまり, $\bm{y}^{ref}$は道具を把持した際の初期の値$\bm{y}^{init}$となる.
  例えば, ハンマーで打撃動作を行うと, 徐々に接触状態が変化し, 持ち方が変わってしまうことがあるが, これを抑制するような制御である.
  この場合, 損失関数は以下のように設計する.
  \begin{align}
    &h_{loss}(\bm{y}^{est}_{seq}, \bm{y}^{ref}_{seq}) = h_{loss, 1}(\bm{y}^{est}_{seq}, \bm{y}^{ref}_{seq}) + h_{loss, 2}(\bm{y}^{est}_{seq}, \bm{y}^{ref}_{seq})\\
    &h_{loss, 1}(\bm{y}^{est}_{seq}, \bm{y}^{ref}_{seq}) = ||\bm{w}_{1}(\bm{f}^{est}_{all, seq}-\bm{f}^{ref}_{all, seq})||_{2}\\
    &w_{1}[i] = \begin{cases}
      1.0\;\;\;(\bm{f}^{est}_{all, seq}[i] \geq \bm{f}^{ref}_{all, seq}[i])\nonumber\\
      \beta\;\;\;\;\;\;(otherwise)\nonumber
    \end{cases}\\
    &h_{loss, 2}(\bm{y}^{est}_{seq}, \bm{y}^{ref}_{seq}) = w_{2}||\bm{\theta}^{est}_{seq}-\bm{\theta}^{ref}_{seq}||_{2} + w_{3}||\bm{l}^{est}_{seq}-\bm{l}^{ref}_{seq}||_{2}
  \end{align}
  ここで, $\bm{f}^{\{est, ref\}}_{all, seq}$は$\bm{y}^{\{est, ref\}_{seq}}$から$\bm{f}$と$\bm{f}_{contact}$を抜き出して縦に並べたもの, $\{\bm{\theta}, \bm{l}\}^{\{est, ref\}}_{seq}$は同様に$\bm{y}$から$\{\bm{\theta}, \bm{l}\}$を抜き出したものを表す.
  また, $w_{\{2, 3\}}$は重み, $w_{1}$は重みのベクトル, $w_{1}[i]$はその$i$番目の要素, $\beta$は定数を表す.
  この$w_{1}$の設計は, 接触センサの性質を考慮したものである.
  接触センサや筋張力センサの値は, マイナス方向には0までしか変化しないが, プラス方向には定格まで大きく変化する.
  つまり, $\beta=1.0$のとき, 最適化の結果として, 接触がマイナス方向に変化しやすくなってしまい, 接触が維持できない.
  そこで, $\beta$を1.0以上に設定することで, 接触を常に保ち続けるような制御指令を生成する.
  また同時に, 関節角度, 筋長に対しても制限をかけることで, 接触を維持しつつも, 筋長・関節角度が大きく変化しないような制御指令を得ることができる.
  ただ, 初期状態を保つような把持安定化ではなく, 狙った接触力を出したいような場合に置いては, $\beta=1.0$が好ましい.

  本研究では, $N_{control}=8$, $C_{batch, control}=4$, $C_{batch, epoch}=3$, $\beta=3.0$, $w_{2}=1.0$, $w_{3}=1.0$とする.
}%

\section{Experiments} \label{sec:experiment}

\begin{figure}[t]
  \centering
  \includegraphics[width=0.6\columnwidth]{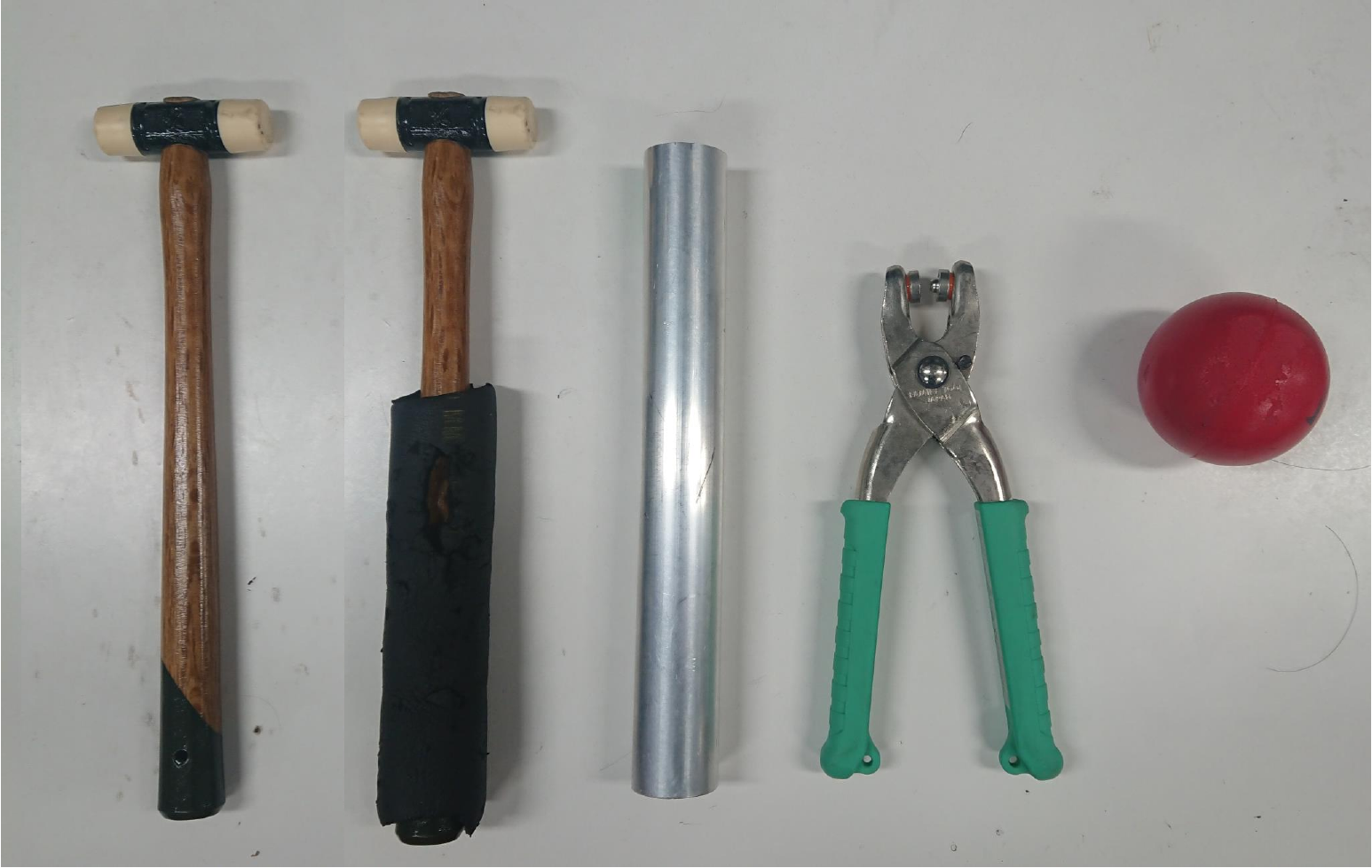}
  \caption{Grasped objects in this study: Hammer, Hammer-S (hammer with soft cover), Cylinder, Gripper, and Ball.}
  \label{figure:used-objects}
  \vspace{-1.0ex}
\end{figure}

\begin{figure}[t]
  \centering
  \includegraphics[width=1.0\columnwidth]{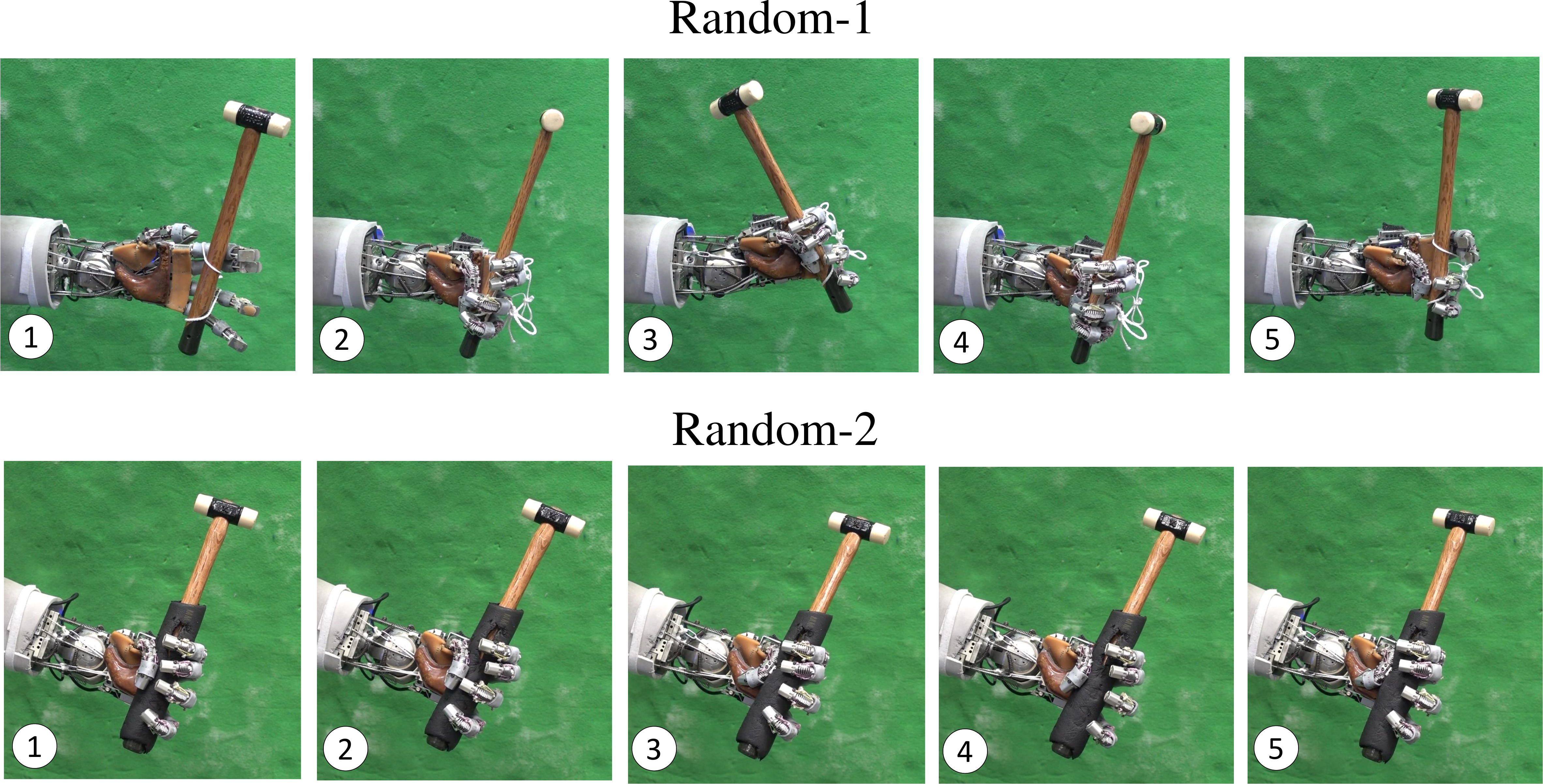}
  \caption{Motion Sequences of the data collection phase of Random-1 and Random-2.}
  \label{figure:data-collection}
  \vspace{-3.0ex}
\end{figure}

\begin{figure}[t]
  \centering
  \includegraphics[width=0.8\columnwidth]{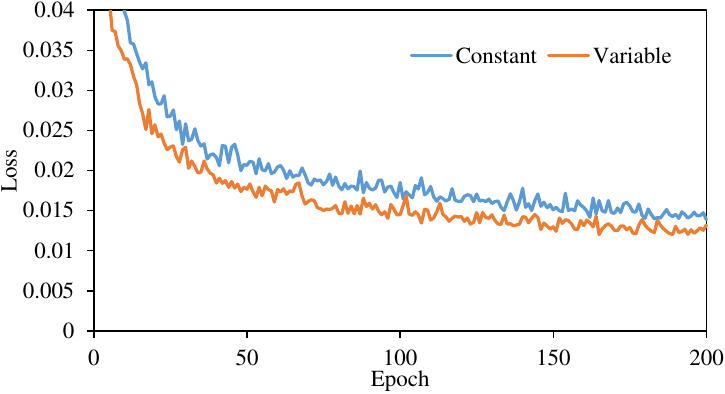}
  \caption{Comparison of the training results of HADYNET between with fixed parametric bias (Constant) and with variable parametric bias (Variable, this study).}
  \label{figure:training-result}
  \vspace{-1.0ex}
\end{figure}

\subsection{Training of HADYNET} \label{subsec:training-exp}
\switchlanguage%
{%
  The grasped objects used in this study are Hammer, Hammer-S (hammer with soft cover), Cylinder, Gripper, and Ball, as shown in \figref{figure:used-objects}.
  We handle the six objects, including None without grasping anything.
  In this study, we conduct two random motions of Random-1 and Random-2 to collect training data.
  Random-1 is a repeating motion that sends random target joint angles over random intervals by converting the target joint angle to the target muscle length using a geometric model as in \figref{figure:musculoskeletal-hand}.
  Random-2 is a motion randomly changing finger muscle lengths $\bm{l}_{finger}$ while grasping the object.
  Especially, Random-2 is important for the contact control explained in \secref{subsec:control}, and consecutive contact changes by different grasps can be obtained.

  Holding these six objects by the hand, Random-1 and Random-2 were each conducted 500 time steps (100 seconds).
  The data collection experiment is shown in \figref{figure:data-collection}.
  Although we must lightly tie up the object to the hand by wires or tapes so as not to drop it regarding Random-1, regarding Random-2, such a deal is unnecessary because the grasping shape of the hand does not largely change.
  Random-1 and Random-2 were each conducted three times regarding one object while changing the initialization, and 36 experiments were conducted with all the objects.
  The number of the collected data was about 18000, and we trained HADYNET using it.
  We compare the loss transition when fixing PB and training PB implicitly as different values (this study) in \figref{figure:training-result}.
  The loss of the latter was smaller by about 20 \%.
  However, because many data without grasping, which is the same data with None, was included regarding each grasped object, the actual difference is considered to be larger.
  Thus, by adding parametric bias, the network can implicitly train the difference of the grasped objects and initializations.
  In this study, the trained network can cope with the difference of the grasped objects and initializations, and by collecting data after a while (e.g. one month), the network becomes able to cope with deterioration over time.
  The deterioration over time can mainly affect the elongation of muscles and the change in the original length of the spring, and this is about the same degree of difference that can be seen due to the irreproducibility of initialization.
  Therefore, we consider that deterioration over time can be taken into account if irreproducibility of initialization can be taken into account.
  Also, it is difficult to determine whether the difference in PBs is due to irreproducibility of initialization or deterioration over time, and so we do not perform any direct experiments on deterioration over time.
}%
{%
  本研究で用いる把持物体は, \figref{figure:used-objects}に示すように, Hammer, Hammer-S (hammer with soft cover), Cylinder, Gripper, and Ballの5つを用いる.
  また, 何も把持しない状態としてNoneを加え, 全部で6つの把持物体について扱う.

  これら6つの把持物体を持たせ, \secref{subsec:training}のRandom-1, Random-2の動作をそれぞれ約500ステップ(100秒)行う.
  その様子を\figref{figure:data-collection}に示す.
  Random-1では把持物体が手から落ちないように紐やテープで軽く括りつけているが, Random-2では把持状態から大きく動作しないため必要ない.
  これを, ハンドのセットアップをやり直しながら3回行い, 一つの把持物体で計6回, 全把持物体で36回の試行を行った.
  集まったデータは計約18000個となり, これを用いてHADYNETを学習させた.
  パラメトリックバイアスを固定して学習させた場合と, タスクごとに別の値をimplicitに学習させた場合(this study)のlossの遷移の比較を\figref{figure:training-result}に示す.
  前者と後者にはlossの下がり方に差があり, 後者の方が約20\%lossが小さいことがわかる.
  ただし, 何も掴んでいない瞬間のNoneと同じデータがそれぞれの把持物体について多く含まれるため, そこまで大きな変化ではない.
  つまり, パラメトリックバイアスを入れることで, 把持物体の違いやキャリブレーションの違いを暗黙的に学習させることができる.
  本実験ではキャリブレーション・把持物体という2つの違いにのみ対応しているが, これを時間間隔を空けて繰り返すことで, 経年劣化等にも対応できるようになる.
  経年劣化は主に筋の伸びやバネの癖がつき初期点が変わってしまう等の影響を持つが, これらは十分にキャリブレーションの違いにも現れうる違いであるため, キャリブレーションの違いを扱うことができれば, 経年劣化も扱うことができると考える.
  また, PBの違いが経年劣化とキャリブレーションどちらによる違いかを区別することは難しく, 本研究では直接的には経年劣化に関する実験は行わない.
}%

\begin{figure}[t]
  \centering
  \includegraphics[width=0.8\columnwidth]{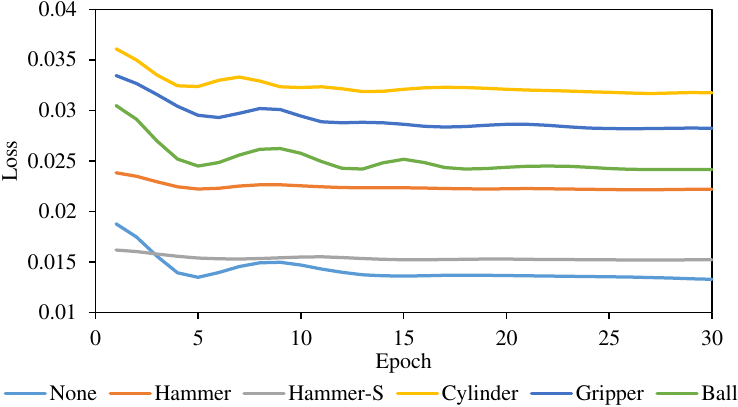}
  \caption{Transition of loss when updating only parametric bias.}
  \label{figure:parametric-bias}
  \vspace{-3.0ex}
\end{figure}

\begin{figure}[t]
  \centering
  \includegraphics[width=1.0\columnwidth]{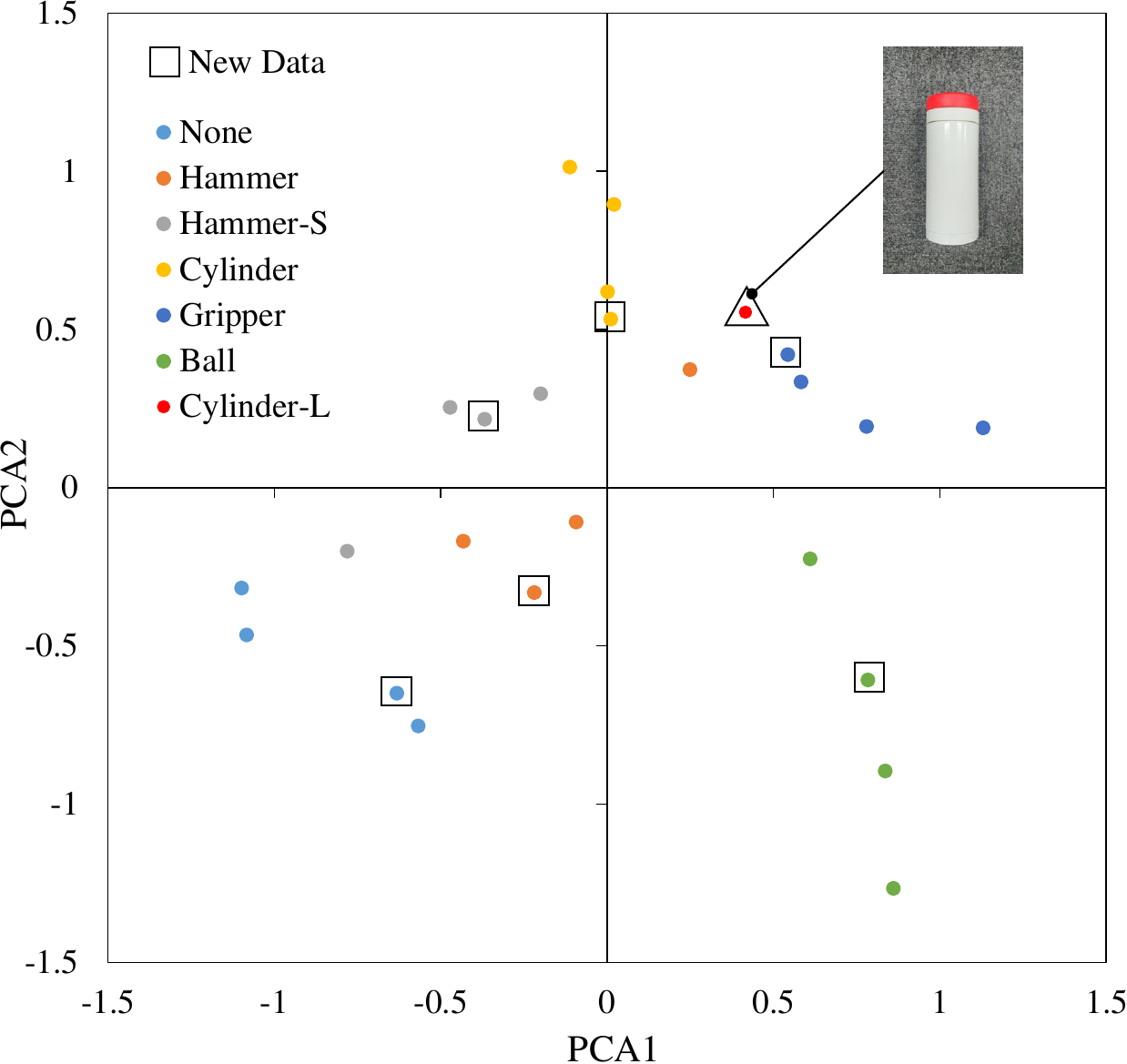}
  \caption{Results of the object recognition. The marks surrounded by squares are new data to recognize what the hand is grasping. The mark surrounded by a triangle is a new object, the large cylinder (Cylinder-L).}
  \label{figure:object-recognition}
  \vspace{-3.0ex}
\end{figure}

\begin{figure}[t]
  \centering
  \includegraphics[width=1.0\columnwidth]{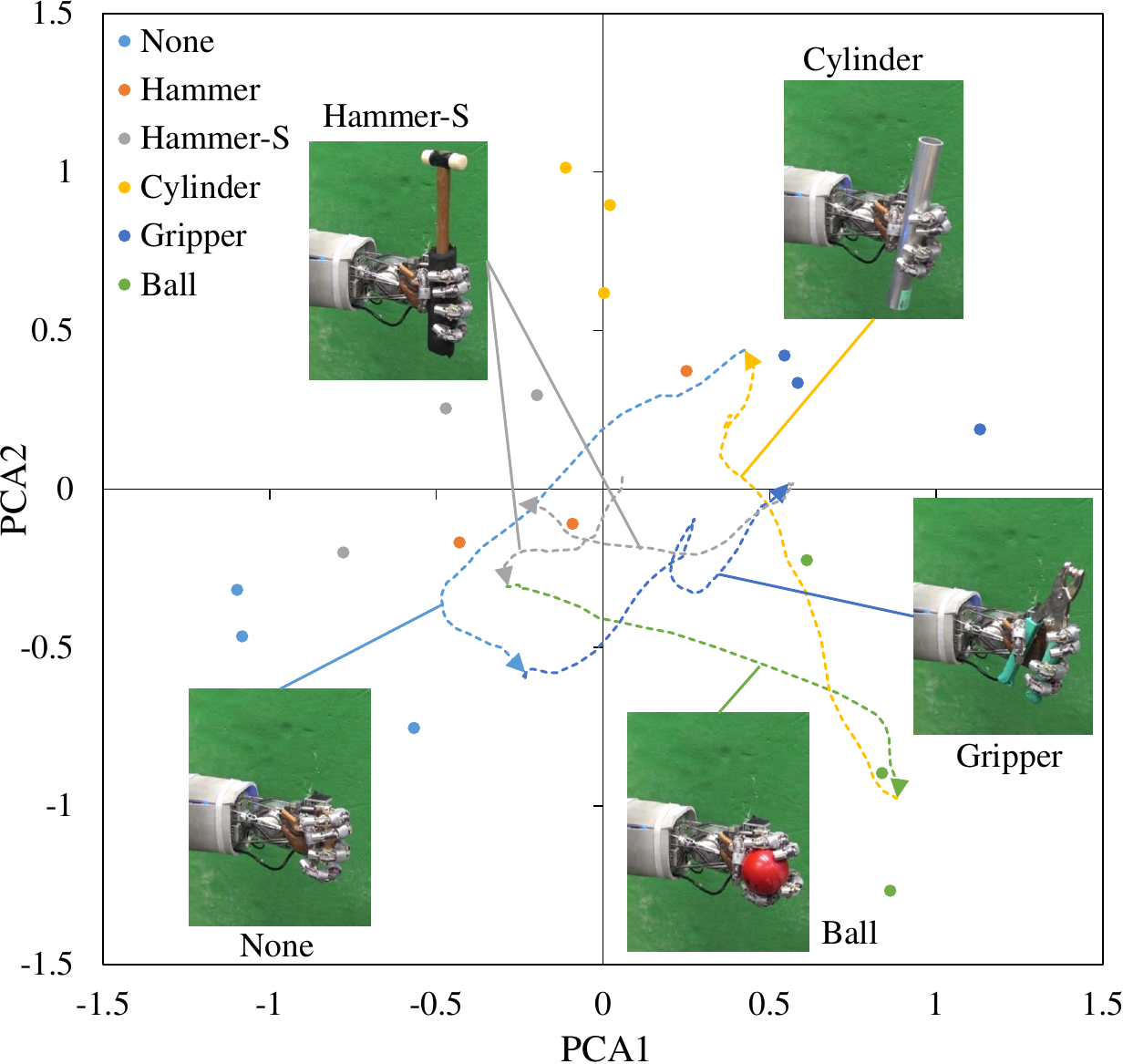}
  \caption{Online update experiment of parametric bias. The graph shows the transition of parametric bias when updating it online.}
  \label{figure:online-pb}
  \vspace{-1.0ex}
\end{figure}

\subsection{Recognition of Grasped Objects} \label{subsec:recognition-exp}
\switchlanguage%
{%
  First, we conducted an experiment on the update of PB.
  We initialized the hand and collected new data regarding six objects.
  We show the loss transition when updating only PB in \figref{figure:parametric-bias}.
  Regarding all the grasped objects, the loss decreased by about 6--28\% by updating only PB.
  Thus, by changing PB, the hand dynamics states with various grasped objects and initializations can be reproduced.

  In \figref{figure:object-recognition}, we plot PBs trained in \secref{subsec:training-exp} and PBs updated using the newly collected data in the two-dimensional space as explained in \secref{subsec:recognition}.
  Each PB was distinctly grouped according to the names of grasped objects.
  The data surrounded by a square is a new data, and regarding each new data, the correct name of the new grasped object can be recognized with the nearest neighbor method.
  Thus, by grasping the target object and moving the hand randomly, the robot can recognize the object without visual information.

  We collected data regarding a new object, Cylinder-L, and updated PB.
  The diameter of Cylinder-L is 68 mm, while the diameter of Cylinder is 32 mm.
  Also, the width of Gripper, which is the largest among objects used for the training, is 60 mm.
  We show the updated PB of Cylinder-L surrounded by a triangle in \figref{figure:object-recognition}.
  This PB is almost at the halfway point of PBs of the Cylinder and Gripper.
  This is considered to be valid because Cylinder-L has the shape characteristics of Cylinder and the size characteristics of Gripper.
  Thus, regarding such a new object, by referring to the trained parametric bias, the robot can recognize the characteristics of the grasped object.

  Finally, we tried an online learning experiment of PB explained in the latter half of \secref{subsec:update}.
  We executed only Random-2 in \secref{subsec:training} over about one and a half minutes each with objects Hammer-S, Ball, Cylinder, None, Gripper, and Hammer-S, in order.
  We show the transition of PB plotted in the two-dimensional space while updating PB online, in \figref{figure:online-pb}.
  The PBs moved in the direction of the arrows, and the transitions of the updated PBs are shown in the same colors as the PBs trained in \secref{subsec:training-exp}.
  From the figure, although the accuracy decreased compared with \figref{figure:object-recognition}, we can see that the current PB moved in the direction of PBs of the current grasped object.
  While PB when grasping Ball moved accurately, regarding the other objects, although the current PBs moved near PBs of the grasped objects, they could not completely reach the correct values.
  This is considered to be because only the data up to $N^{online}_{max}$ is used, the data used to update PB always changes, and the loss does not decrease completely.
  However, by updating parametric bias online, we can see that the robot can gradually obtain the characteristics of the grasped object.
}%
{%
  ハンドのキャリブレーションを行い新しく6つの把持物体のデータを得た.
  これを用いてParametric Biasのみの更新を行った際のlossの変化を\figref{figure:parametric-bias}に示す.
  全ての把持物体について, Parametric Biasのみを更新することでlossが6--28\%下がっていることがわかる.
  つまり, Parametric Biasを変更することで, 様々な把持物体・キャリブレーションを再現することができている.

  \secref{subsec:training-exp}で得られたParametric Biasと, 新しいデータによって得られたParametric Biasを, \secref{subsec:recognition}で説明したように2次元平面上にプロットする(\figref{figure:object-recognition}).
  それぞれの把持物体のParametric Biasが, 上手くグルーピングされていることがわかる.
  四角で囲まれたデータが新しいデータであるが, それぞれのデータについてnearest neighborによって把持物体を正しく識別することができている.
  つまり, 対象物体を把持し, それをランダムに握ったり離したりすることで, その物体が何であるかを, 視覚を使わずに判断することができる.

  また, 新しい物体Cylinder-Lに関してもデータを取り, Parametric Biasを学習させた.
  ここで, Cylinderの直径は32 mm, Cylinder-Lの直径は68 mmである.
  また, trainingに使用した物体で最も把持部分が大きいGripperは, 把持部分の幅が60 mmである.
  \figref{figure:object-recognition}に獲得されたCylinder-LのParametric Biasの値(Triangleに囲まれている)を示すが, これはCylinderとGripperのクラスの中間に存在している.
  Cylinderの円筒という形の特徴とGripperの大きさの特徴を併せ持ったCylinder-Lの位置としては妥当に思える.
  つまり, このように新しい物体に関しても, trainingのデータを参照することで, 物体の特徴を掴むことができる.

  最後に, \secref{subsec:update}の後半で説明した, Parametric Biasをオンラインで更新することを試みる.
  Hammer-S, Ball, Cylinder, None, Gripper, Hammer-Sの順で物体を約1分半ずつ握って\secref{subsec:training}のRandom-2を実行する.
  この間常にオンライン更新を実行しておき, その際のParametric Biasを2次元平面上に投影したときの遷移を\figref{figure:online-pb}に示す.
  矢印の方向にParametric Biasは移動しており, training時に得られたParametric Biasのプロットの色と遷移の線の色は一致している.
  図から, \figref{figure:object-recognition}のように一度に学習させる場合に比べると精度は落ちるが, 現在のParametric Biasが, 今把持している物体のクラスの方向へ動いていることがわかる.
  特に, BallはParametric Biasが上手く遷移しているのに対して, その他は, 現在把持物体のクラスの近傍まではいくものの, 最後まで近づききれていないような挙動をしている.
  これは, 最大で$N^{online}_{max}$までのデータしか用いず, 常にデータが変わっていくため, lossが下がりきっていないためだと思われる.
  しかし, このようにオンラインでParametric Biasを更新することで, 徐々に把持物体の傾向を掴むことができることがわかった.
}%

\begin{figure}[t]
  \centering
  \includegraphics[width=1.0\columnwidth]{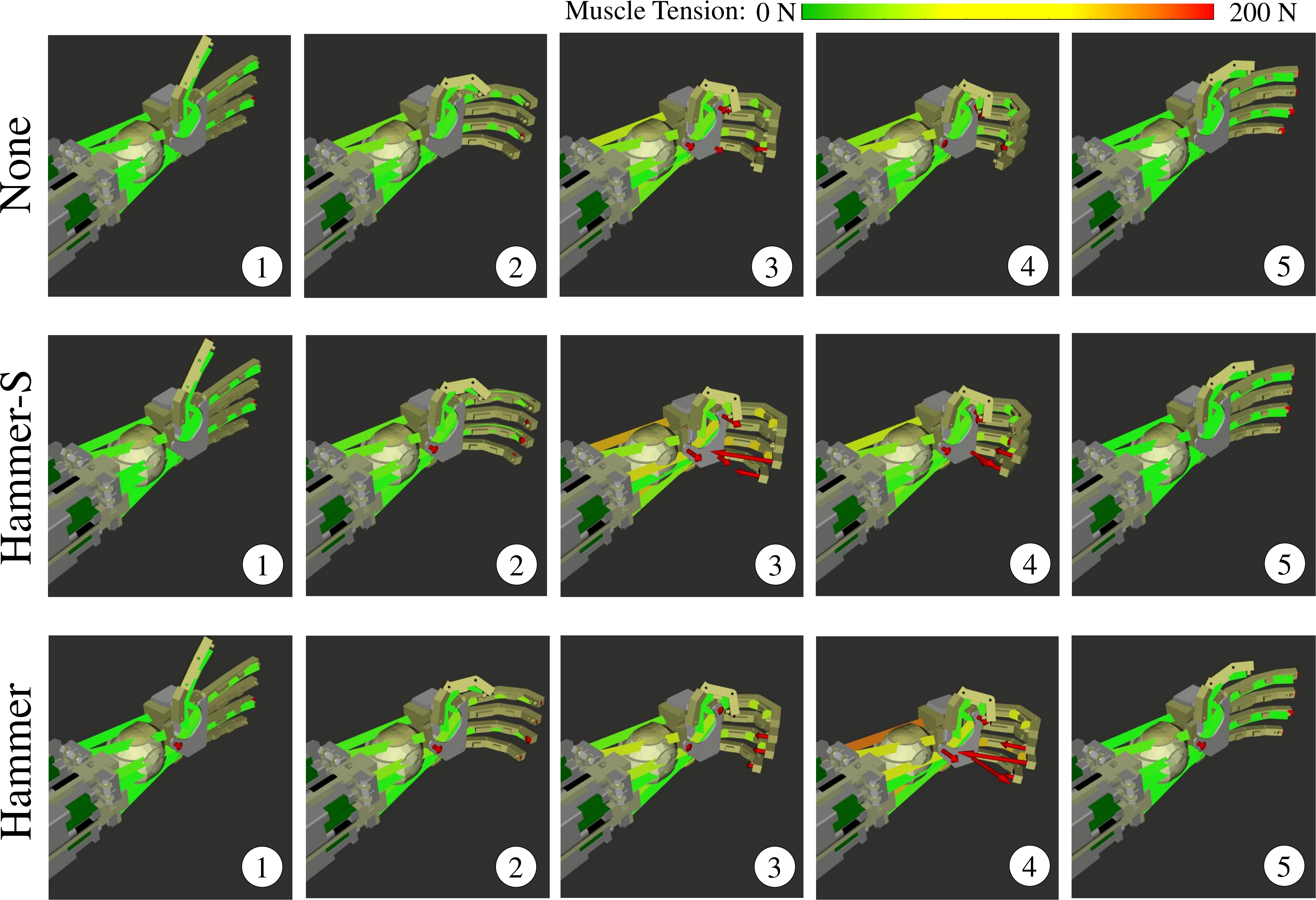}
  \caption{Contact simulation using parametric biases of None Hammer-S, and Hammer.}
  \label{figure:contact-simulation}
  \vspace{-3.0ex}
\end{figure}

\subsection{Contact Simulation} \label{subsec:simulation-exp}
\switchlanguage%
{%
  We show the simulation results of \secref{subsec:simulation} with None, Hammer-S, and Hammer in \figref{figure:contact-simulation}.
  Regarding each, PBs trained in \secref{subsec:training-exp} were used.
  Here, the longer the arrow is, the larger the contact force is, and the redder the muscle color is, the higher the muscle tension is.
  The joint angles of fingers are calculated by using a geometric model and Kalman Filter.
  Regarding None, contact force and muscle tension did not largely change when grasping.
  When grasping Hammer-S, large contact forces were exerted at the loadcells of the fingertips and the palm.
  Also, the muscle tensions were high and some muscle colors became orange.
  Comparing Hammer-S and Hammer, the contact force and muscle tension of Hammer-S increased at an early stage of grasping because the diameter of Hammer-S is larger than that of Hammer due to the soft cover.
  Also, because of the lack of the soft cover, the impact of the grasping is not absorbed regarding Hammer.
  Therefore, in Hammer, it is observed that higher contact forces are suddenly exerted than in Hammer-S at the moment of contact.
  Thus, we can change the hand dynamics only by changing PB.
  This can be used for reinforcement learning of soft robotic hands which are difficult to modelize.
}%
{%
  HADYNETを使った, \secref{subsec:simulation}におけるsimulationの様子を, None, Hammer-S, Hammerについて\figref{figure:contact-simulation}に示す.
  それぞれにおいて, \secref{subsec:training-exp}で得られたParametric Biasをセットしている.
  ここでは, 矢印が大きいほど接触力が大きく, 筋の色が緑から赤くなるほど筋張力が高いことを表している.
  Noneの場合は接触力・筋張力ともに大きな変化は現れていない.
  それに対して, Hammer-Sの場合は指先・手のひらのロードセルに大きな力が加わり, 筋張力も高まっていることがわかる.
  Hammer-SとHammerを比べると, Hammer-Sはsoft coverによって径がHammerよりも大きくなっているため, 把持の早い段階で接触力・筋張力の高まりが見られる.
  Hammerはsoft coverがないため把持の際の衝撃が吸収できず, 接触が生じた瞬間に突然Hammer-Sよりも大きな力が出ていることもわかる.
  このように, Parametric Biasの変化のみでハンドのダイナミクスを変化させることが可能となった.
  これは, モデリング困難な柔軟ハンドのシミュレーション上での強化学習に用いることができると考える.
}%

\begin{figure}[t]
  \centering
  \includegraphics[width=0.9\columnwidth]{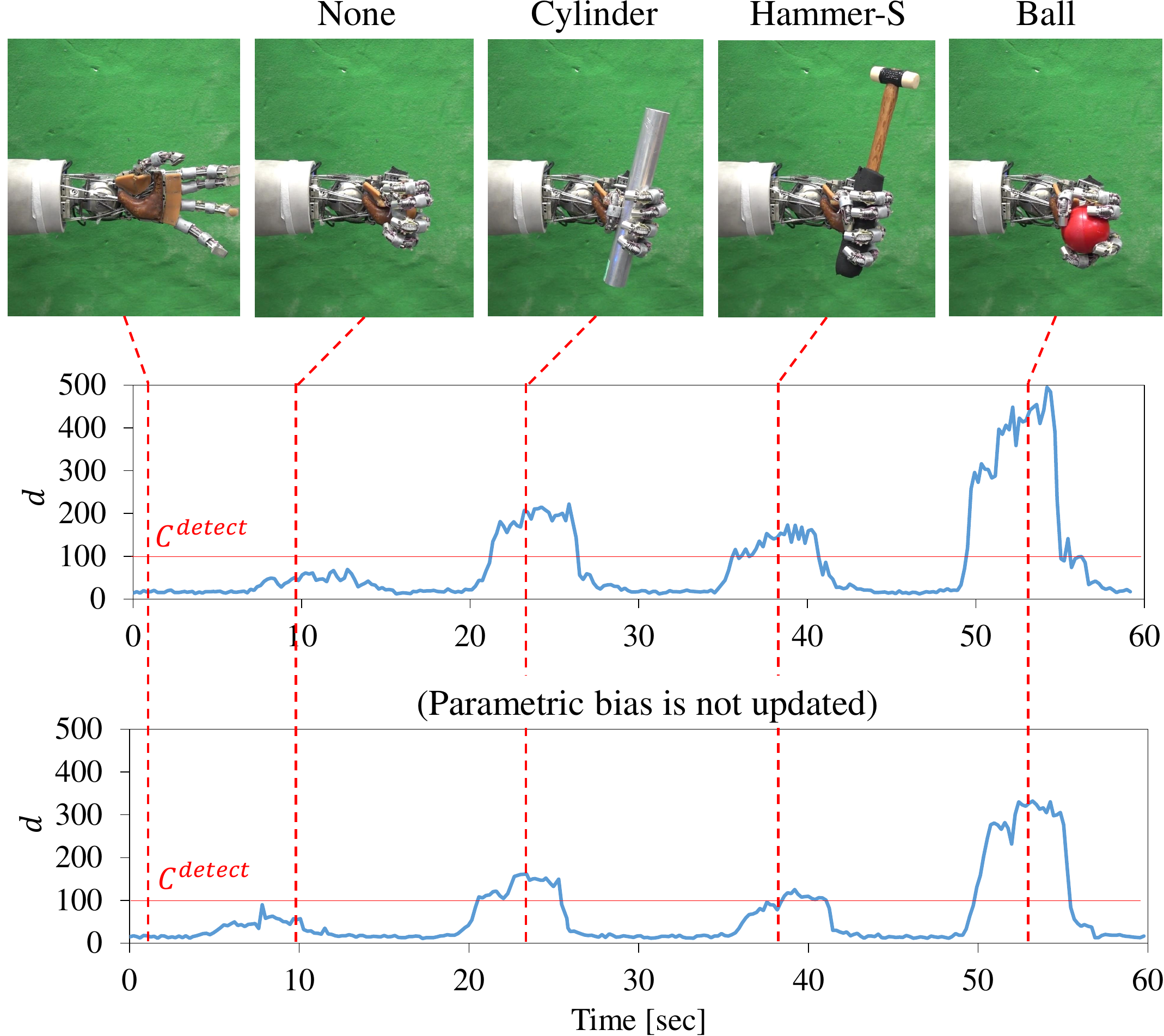}
  \caption{Contact detection experiment when grasping various objects. The middle graph shows $d$ when updating parametric bias and the lower shows $d$ when using parametric bias trained previously.}
  \label{figure:contact-detection}
  \vspace{-1.0ex}
\end{figure}

\begin{figure}[t]
  \centering
  \includegraphics[width=1.0\columnwidth]{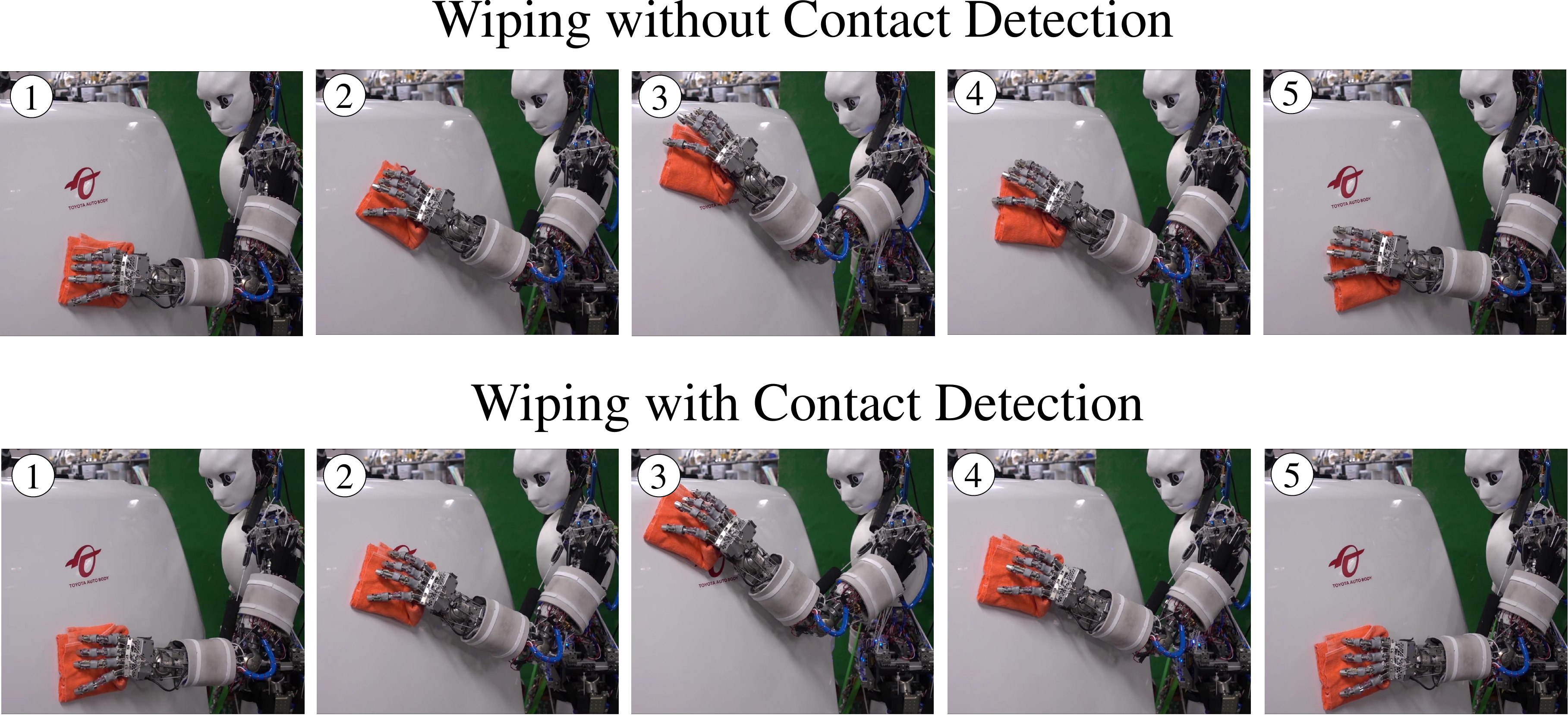}
  \caption{Wiping experiments with and without contact detection.}
  \label{figure:clearning-experiment}
  \vspace{-3.0ex}
\end{figure}

\begin{figure}[t]
  \centering
  \includegraphics[width=0.95\columnwidth]{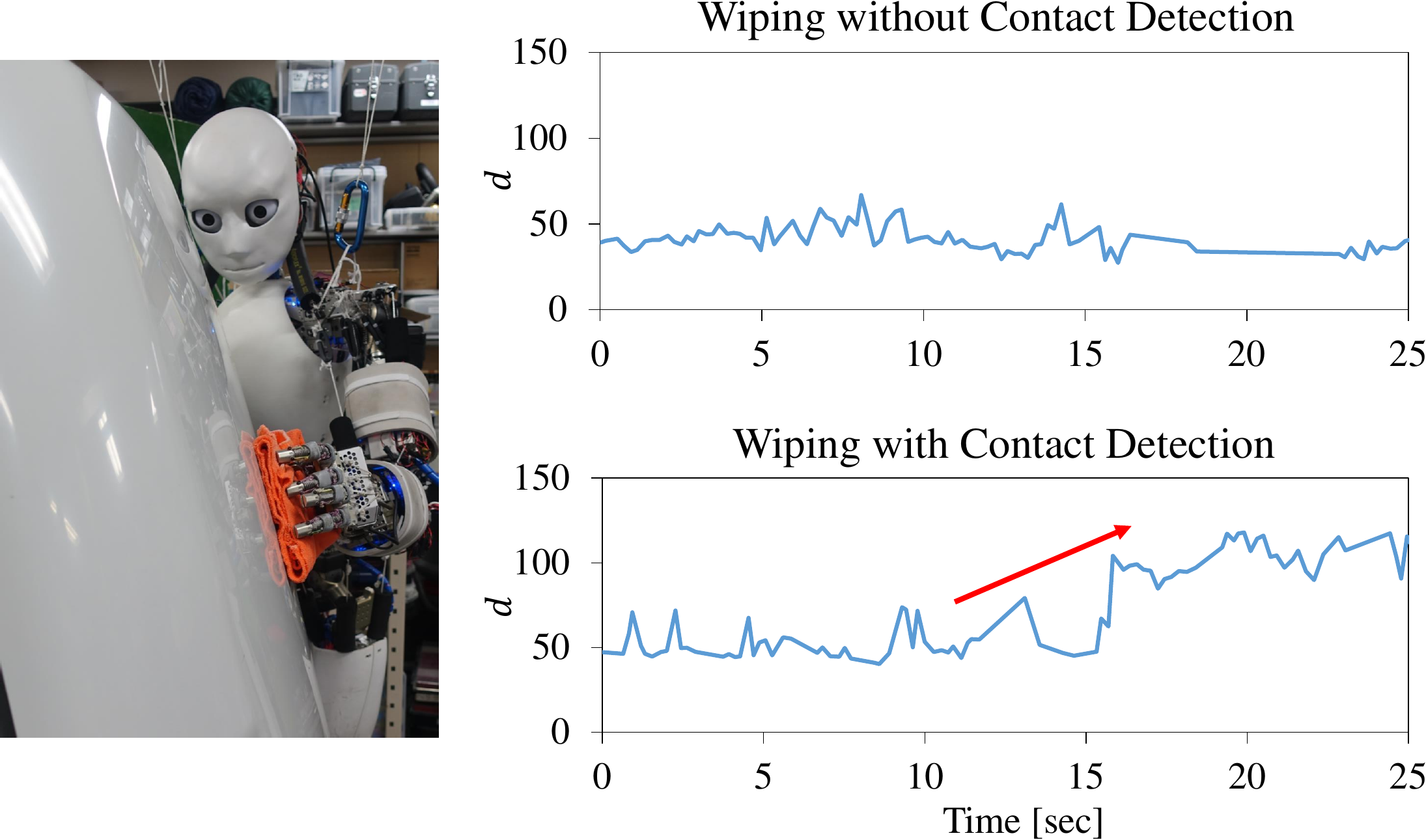}
  \caption{The transitions of $d$ when wiping with and without contact detection. The left figure shows the inclination of the target surface to be wiped.}
  \label{figure:clearning-graph}
  \vspace{-3.0ex}
\end{figure}

\subsection{Contact Detection} \label{subsec:detection-exp}
\switchlanguage%
{%
  First, we observed how the value $d$ in \secref{subsec:detection} for contact detection changes when grasping various objects.
  We show the transition of $d$ when updating PB at the current hand dynamics state of None (without grasping anything) and when grasping None, Cylinder, Hammer-S, and Ball, in order, in the middle graph of \figref{figure:contact-detection}.
  In this study, because $C^{detect}=100$, we can see correct transitions in that None was not detected and Cylinder, Hammer-S, and Ball were detected.
  When referring to \figref{figure:object-recognition}, Ball and Cylinder, whose PBs are distant from None, showed larger $d$ than Hammer-S, whose PB is near None.
  On the other hand, we show the result of contact detection when using PB of None trained in \secref{subsec:training-exp} without updating PB at the current state of initialization, in the lower graph of \figref{figure:contact-detection}.
  $d=89$, which is near $C^{detect}$, was measured regarding None, and $d=114$, which is near $C^{detect}$, was measured regarding Hammer-S at most.
  This causes the contact to be detected on and off due to minor errors.
  Thus, PB includes the hand dynamics information of initialization, and so we should update PB again when the hand is initialized, even if the grasped object is the same.

  Second, the contact detection is used for not only detecting but also keeping appropriate contact.
  We conducted an experiment of wiping a surface while keeping $d$ at $C^{detect}$.
  In detail, the robot moves the hand away from the surface when contact is detected and moves toward the surface when contact is not detected.
  We show the cleaning experiments with and without contact detection in \figref{figure:clearning-experiment}, and show the transitions of $d$ in \figref{figure:clearning-graph}.
  In \figref{figure:clearning-experiment}, while the cloth slipped from the hand without contact detection, the surface was traced well with contact detection.
  As shown in the left figure of \figref{figure:clearning-graph}, the surface to be cleaned has an inclined structure in which the hand leaves the surface as it goes up.
  Thus, the hand moved straight up without contact detection, the hand left the surface, and the cloth slipped from the hand.
  Without contact detection, $d$ did not largely change.
  Compared with this case, when using contact detection, the hand not only moves straight up but also gets close to the surface, and the hand can move along the surface.
  When moving the hand up, because the inclination of the surface and the trajectory of the hand getting close to the surface were almost the same, we could not see a large change in $d$.
  However, when moving the hand down, as the hand and surface got closer, $d$ became large, and $d$ finally reached near $C^{detect}$.
  Thus, we can make use of the contact detection passively and actively.
}%
{%
  まず, 様々な物体を握らせた時に, 接触検知の値$d$がどのような挙動をするのかを観察する.
  現在のキャリブレーション状態で一度Parametric BiasをNoneに適合させ, None, Cylinder, Hammer-S, Ballを持たせたときの$d$の遷移を\figref{figure:contact-detection}の中図に示す.
  本研究では$C^{detect}=100$であり, Noneは検知されず, Cylinder, Hamer-S, Ballを握ったときに正しく接触が検知出来ていることがわかる.
  \figref{figure:object-recognition}と照らし合わせると, Noneから値が遠いBallやCylinderの方が, Noneから近いHammer-Sに比べて$d$が大きく出ていることがわかる.
  これに対して, training時に得られたNoneのParametric Biasを用いて接触検知を行った結果を\figref{figure:contact-detection}の下図に示す.
  物体を把持したときの$d$にメリハリがなくなり, Noneのときは89という$C^{detect}$に近い値が, Hammer-Sのときも最大で114という$C^{detect}$に近い値が計測された.
  これでは多少の誤差で接触が検知されたりされなかったり, という現象が起きてしまう.
  つまり, Parametric Biasはcalibrationの状態を含み, もしハンドをキャリブレーションし直したなら, 把持物体が同じであっても一度Parametric Biasを更新することが望ましい.

  次に, 接触を検知するだけでなく, 接触を適切に保つような動作も可能である.
  接触を保つ, つまり$d$が$C^{detect}$になるように制御しつつ, 布巾で掃除をするタスクを行った.
  具体的には, 接触が検知されたら接触面から離れる方向に, 検知されなければ接触面に押し付ける方向に逆運動学を解いて少しずつ動かしていく.
  \figref{figure:clearning-experiment}に接触保持を入れない場合と入れた場合の掃除の様子を, \figref{figure:clearning-graph}にその際の$d$の遷移を示す.
  \figref{figure:clearning-experiment}では, 接触保持を入れない場合は布巾が手からズレ落ちてしまっているのに対して, 接触保持を入れた場合はしっかりと面をなぞれていることがわかる.
  \figref{figure:clearning-graph}の左図に示すように, この掃除する面は上に行くほど離れるような斜めの構造になっている.
  つまり, 接触保持を入れない場合は手が直上に動き, 徐々に接触面から手が離れて布巾がズレてしまっているのである.
  その際の$d$は, 一様でほとんど変化がないことがわかる.
  これに対して, 接触保持を入れた場合は手を直上に動かすだけでなく, 徐々に接触面方向にも動かされるため, 面に沿って動作ができている.
  手を上に動かすときは接触面の傾きと接触保持による動作がほとんど同じなため$d$に大きな変化は見られないが, 手を下に動かす際には逆に接触力が上昇する方向のため, $d$が上昇し, $C^{detect}$付近で変化している.
  よって, 接触検知を能動かつ受動的に活用することができる.
}%

\begin{figure}[t]
  \centering
  \includegraphics[width=1.0\columnwidth]{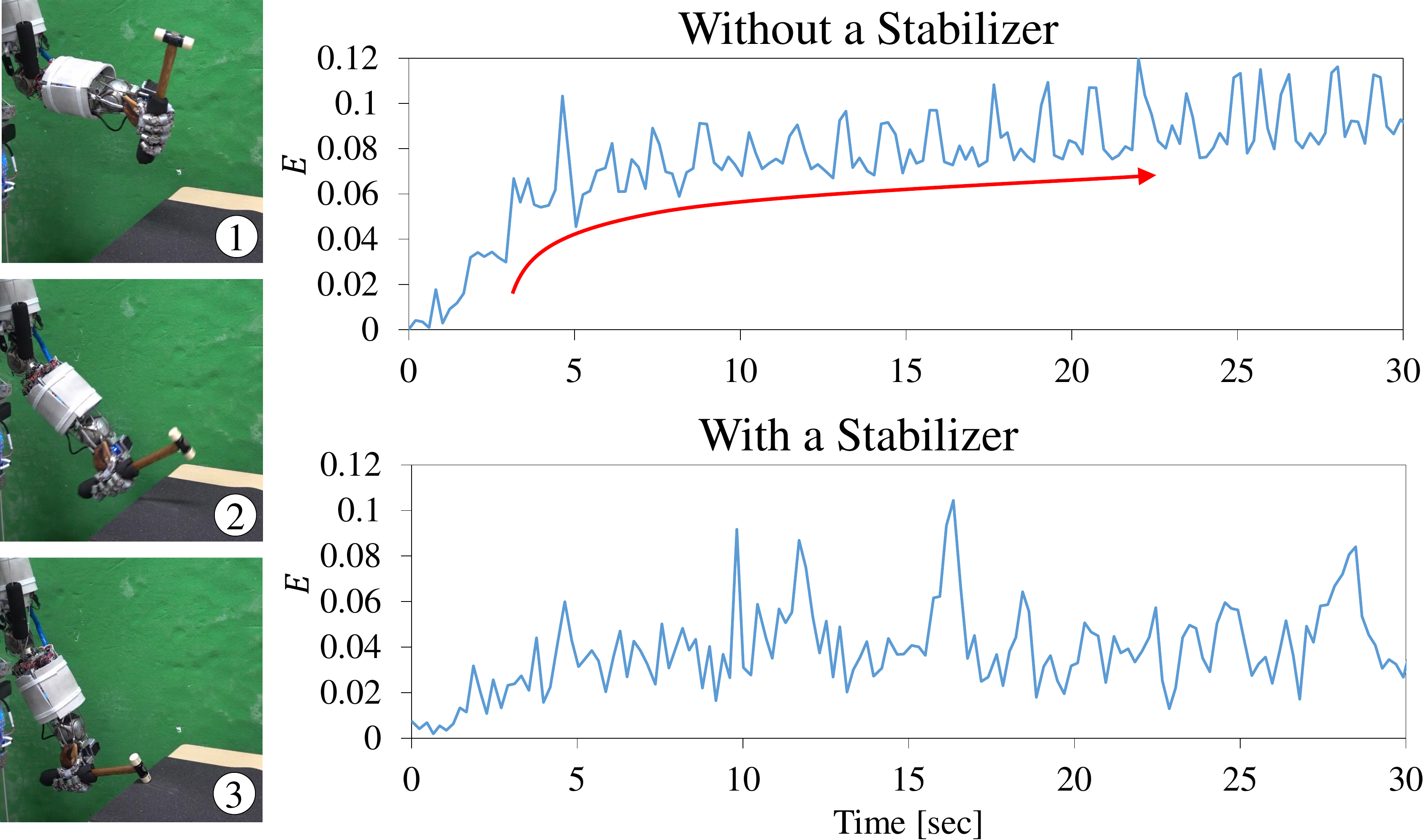}
  \caption{Hammer hitting experiments with and without the grasping stabilizer. The graph shows the transition of the evaluation value $E$.}
  \label{figure:grasping-stabilizer}
  \vspace{-3.0ex}
\end{figure}

\subsection{Contact Control} \label{subsec:control-exp}
\switchlanguage%
{%
  We applied the strategy of grasping stabilizer explained in \secref{subsec:control} to the hammer hitting operation, and evaluated the contact stability.
  First, we obtained $D'$ regarding Hammer-S, and updated PB.
  Second, regarding the cases with and without the grasping stabilizer, we verified the transition of evaluation value $E$ (explained subsequently), when hitting a table using Hammer-S over 30 seconds as shown in \figref{figure:grasping-stabilizer}.
  $\bm{y}^{ref}$ is $\bm{y}^{init}$, which is the initial value of $\bm{y}$ when grasping Hammer-S.
  In this study, because the main purpose is keeping the initial contact state, we use $h_{loss, 1}$ in \secref{subsec:control} as $E$.
  The smaller the $E$, the more the grasping is stabilized.
  We show the experimental results in \figref{figure:grasping-stabilizer}.
  In the case without grasping stabilizer, $E$ gradually became large, and the initial contact state collapsed.
  Compared with this case, when using grasping stabilizer, $E$ did not largely change from near 0.3, and even when $E$ became large, the value reverted rapidly.
  Thus, by using HADYNET, the hand can realize the target contact state.
}%
{%
  \secref{subsec:control}で説明した把持安定化戦略を, ハンマー操作の際に適用し, 接触の安定化が可能かどうかを評価する.
  まず, 使用するHammer-Sに関して$D'$を取得してParametric Biasを更新する.
  次に, 把持安定化戦略を使用した場合と使用しない場合について, \figref{figure:grasping-stabilizer}の上図のようにハンマーで机を叩く動作を30秒間行い, その際の評価値$E$の遷移を確認する.
  $\bm{y}^{ref}$は最初にHammer-Sを握ったときの値$\bm{y}^{init}$とする.
  本研究では道具把持初期の接触を常に保つような安定化を行うため, $E$は\secref{subsec:control}における$h_{loss, 1}$を用い, これは小さいほうが把持が安定していると言える.
  実験結果を\figref{figure:grasping-stabilizer}の中図・下図に示す.
  把持安定化戦略を用いない場合は, 徐々に$E$が大きくなり, 初期の把持が崩れてしまっていることがわかる.
  これに対して, 把持安定化戦略を用いたときは, 試行を重ねても$E$は0.3付近から大きく変化せず, 大きな値になっても, すぐに元に戻っている.
  つまり, HADYNETを用いることで, 指令した接触状態を実現するような制御が実行可能である.
}%

\section{CONCLUSION} \label{sec:conclusion}
\switchlanguage%
{%
  In this study, for the flexible musculoskeletal hand, we developed a method of the recognition of grasped objects, contact simulation, detection, and control, coping with its deterioration over time, irreproducibility of initialization, and the difference in grasped objects.
  By constructing the sensor state equation using a recurrent neural network with parametric bias, the different dynamics of the hand is implicitly learned.
  The hand can recognize the grasped object by the difference of parametric bias and its contact states can be simulated by the forwarding of the network.
  The hand can conduct the contact detection using the prediction error of the network, and can conduct the contact control by making the current sensor state close to the target value through backpropagation technique to the control input.
  We integrated these various components into one network, and verified the realization of various dynamics by changing only the parametric bias of the network.

  In future works, we would like to focus on the task realization using these components.
}%
{%
  本研究では, 柔軟ハンドの把持物体やキャリブレーションのズレ, 経年劣化等のダイナミクスの変化に対応した, 把持物体認識・接触シミュレーション・接触検知・接触制御の手法を開発した.
  パラメトリックバイアスを用いた再帰的ニューラルネットワークによる状態方程式の構築を行うことで, それらダイナミクスの違いを暗黙的に学習させることが可能となる.
  また, Parametric Biasの違いによる把持物体の認識, ネットワークのフォワーディングによる接触シミュレーションが可能である.
  予測誤差を用いた接触検知, 予測結果を指令値に近づけるような制御指令値を誤差逆伝播法により求める接触制御法を開発した.
  様々なコンポーネントが一つのネットワークで構築可能であり, そのネットワークの一部を更新するのみで様々なダイナミクスを表現できることを確認した.

  今後, これらのコンポーネントを用いたタスク実現に力を入れていきたい.
}%

{
  \bibliographystyle{IEEEtran}
  \bibliography{main}
}

\end{document}